\documentclass{article}

\PassOptionsToPackage{numbers, compress}{natbib}
    \usepackage[preprint]{neurips_2025}

\usepackage[utf8]{inputenc} % allow utf-8 input
\usepackage[T1]{fontenc}    % use 8-bit T1 fonts
\usepackage{hyperref}       % hyperlinks
\usepackage{url}            % simple URL typesetting
\usepackage{booktabs}       % professional-quality tables
\usepackage{amsfonts}       % blackboard math symbols
\usepackage{nicefrac}       % compact symbols for 1/2, etc.
\usepackage{microtype}      % microtypography
\usepackage{xcolor}         % colors

\usepackage{array}
\usepackage[normalem]{ulem}
\usepackage{graphicx}
\usepackage{wrapfig}
\usepackage{subcaption}
\usepackage{multirow}
\usepackage{amsmath}
\usepackage{amssymb}
\usepackage{mathtools}
\usepackage{amsthm}
\usepackage{fontawesome}
\usepackage{algorithm}
\usepackage{algorithmic}

\usepackage[capitalize,noabbrev]{cleveref}

\definecolor{citecol}{HTML}{6F130C}
\definecolor{tableofcontent}{HTML}{1F4A83}
\definecolor{urlcol}{HTML}{2470D8}

\usepackage{hyperref}
\hypersetup{
    colorlinks=true,       % false: boxed links; true: colored links
    linkcolor=tableofcontent,          % color of internal links (change box color with linkbordercolor)
    citecolor=citecol,        % color of links to bibliography
    urlcolor=blue,           % color of external links
}

\theoremstyle{plain}
\newtheorem{theorem}{Theorem}[section]
\newtheorem{proposition}[theorem]{Proposition}
\newtheorem{lemma}[theorem]{Lemma}

\theoremstyle{definition}
\newtheorem{definition}[theorem]{Definition}

\newtheorem{remark}[theorem]{Remark}
\newcommand{\bs}{\hfill $\blacksquare$}

\def\eqref#1{Eq.~\ref{#1}}
\def\tabref#1{Tab.~\ref{#1}}
\def\figref#1{Fig.~\ref{#1}}

\allowdisplaybreaks[3]

\title{How Particle System Theory Enhances Hypergraph Message Passing}
\author{%
  Yixuan Ma \\
  Shanghai Jiao Tong University \\
  \texttt{mayx5901@sjtu.edu.cn} \\
  \And
  Kai Yi \\
  University of Cambridge \\
  \texttt{ky347@cam.ac.uk} \\
  \AND
  Pietro Li{\`o} \\
  University of Cambridge \\
  \texttt{pl219@cam.ac.uk}
  \And
  Shi Jin \\
  Shanghai Jiao Tong University \\
  \texttt{shijin-m@sjtu.edu.cn} \\
  \And
  Yu Guang Wang\thanks{Corresponding author} \\
  Shanghai Jiao Tong University \\
  \texttt{yuguang.wang@sjtu.edu.cn} \\
}

\begin{document}

\maketitle

\begin{abstract}
Hypergraphs effectively model higher-order relationships in natural phenomena, capturing complex interactions beyond pairwise connections. We introduce a novel hypergraph message passing framework inspired by interacting particle systems, where hyperedges act as fields inducing shared node dynamics. By incorporating attraction, repulsion, and Allen-Cahn forcing terms, particles of varying classes and features achieve class-dependent equilibrium, enabling separability through the particle-driven message passing. We investigate both first-order and second-order particle system equations for modeling these dynamics, which mitigate over-smoothing and heterophily thus can capture complete interactions. The more stable second-order system permits deeper message passing. Furthermore, we enhance deterministic message passing with stochastic element to account for interaction uncertainties. We prove theoretically that our approach mitigates over-smoothing by maintaining a positive lower bound on the hypergraph Dirichlet energy during propagation and thus to enable hypergraph message passing to go deep. Empirically, our models demonstrate competitive performance on diverse real-world hypergraph node classification tasks, excelling on both homophilic and heterophilic datasets.
\end{abstract}

%%%%%%%%%%%%%%%%%%%%%%%%%%%%%%%%%%%%%%%%%%%%%%%%%%%%%%%%%%%%%%%%%%%%%%%%%%%%
% Introduction
%%%%%%%%%%%%%%%%%%%%%%%%%%%%%%%%%%%%%%%%%%%%%%%%%%%%%%%%%%%%%%%%%%%%%%%%%%%%
\section{Introduction}\label{Intro}
Hypergraph Neural Networks (HNNs)~\cite{feng2019hypergraph}, built upon Graph Neural Networks (GNNs)~\cite{kipf2016semi}, have demonstrated remarkable success in modeling higher-order relationships involving multiple nodes~\cite{kim_hypergraph}. In complex systems such as social networks~\cite{bazaga2024hyperbert,li2025hypergraph} and biomolecular interactions~\cite{vinas2023hypergraph,wu2024se3set}, hypergraphs can effectively capture complex group dynamics compared to pairwise graphs. 

\begin{figure}[!t]
    \includegraphics[width=\textwidth]{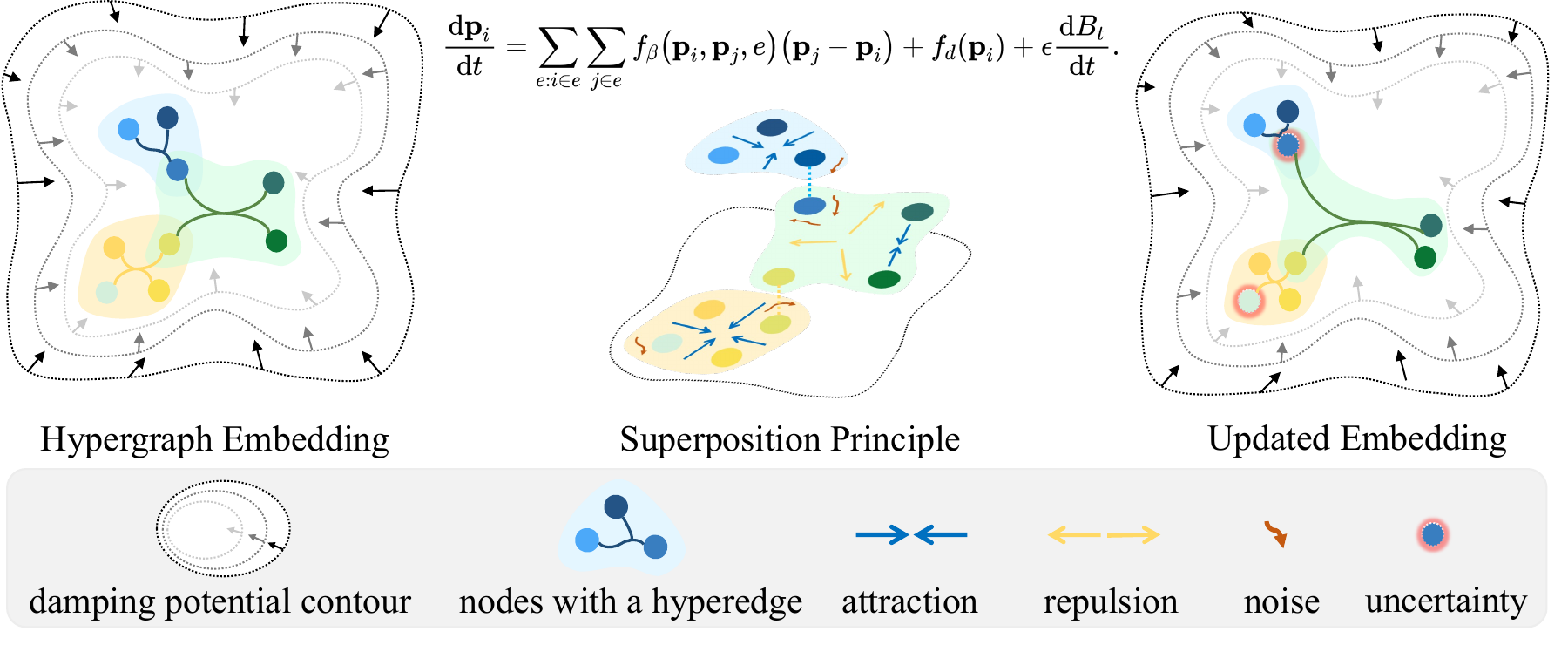}
    \caption{An illustration for \textsf{HAMP} framework. 
    The property $\mathbf{p}$ can account for feature or velocity.
    }
    \label{fig:main_architecture}
\end{figure}

The multi-node, higher-order nature of hypergraphs naturally lends itself to an analogy with particle dynamical systems. This is because hypergraph message passing, much like the inherent interactions in particle motion, fundamentally involves multiple interacting components. Driven by this insight, we introduce Hypergraph Atomic Message Passing (HAMP), a novel framework that reframes hypergraph message passing through the lens of interacting particle systems. Our key innovation in HAMP is the conceptualization of each hyperedge as a dynamic field that governs the shared dynamics of nodes. As Figure~\ref{fig:main_architecture} shows, HAMP updates hypergraph embeddings by superimposing the forces exerted by the hypergraph's nodes (particles). This framework offers significant advantages, including the ability to mitigate over-smoothing and facilitate deep message passing on both homophilic and heterophilic datasets, a claim supported by our experimental findings in Section~\ref{ab_studies}.

Particle theory takes various forms, traditionally categorized into first-order and second-order systems. While first-order systems offer a direct approach to defining evolution, the second-order systems are inherently more stable, converging towards an asymptotically stable state. Both first-order and second-order systems exhibit desirable separability properties stemming from the balance between attractive and repulsive forces~\cite{jin2021collective,kolokolnikov2013emergent}. Repulsive forces are crucial for separating distinct features, while attractive forces encourage particles of the same category to cluster, allowing for the refinement of essential features. 
Through this mechanism, we construct the interaction forces within \textsf{HAMP}.

Inspired by particle systems and informed by experimental observations, we identify a critical need for a delicate balance between attractive and repulsive forces. Excessive attraction often leads to feature over-smoothing~\cite{nt2019revisiting, oono2019graph}, while unchecked repulsion can cause feature explosion~\cite{wang2023acmp}. To address these issues and ensure system stability, we introduce the Allen-Cahn force~\cite{allen1979microscopic} as a balancing mechanism. We theoretically demonstrate that the resulting system exhibits favorable separability and that its solutions converge to an equilibrium state, thereby mitigating both over-smoothing and feature explosion. Our focus on second-order systems is further motivated by their inherent theoretical stability, which ensures robustness even with deeper network layers. 
In practical tasks, ambiguous cases where distinct states are hard to differentiate often lead to uncertainty~\cite{wang2024ugnn}. To address this, we incorporate a stochastic term into \textsf{HAMP}, driven by Brownian motion, which results in a stochastic differential equation. Our main contributions are summarized as follows: 
\begin{itemize}
    \item We propose \textsf{HAMP}, a novel hypergraph message passing framework based on  particle system theory, filling in the theoretical gap in understanding hypergraph message passing from the perspective of particle system. 
    \item We design two hypergraph message passing algorithms, \textsf{HAMP-I} and \textsf{HAMP-II}, constructed through first-order and second-order particle dynamical systems. 
    \item Theoretically, we prove that \textsf{HAMP} maintains a strictly positive lower bound on the hypergraph Dirichlet energy, effectively resisting over-smoothing.
    \item Empirically, through numerical experiments, we should that \textsf{HAMP} achieves competitive results on node classification benchmarks. Notably, on heterophilic hypergraphs, \textsf{HAMP} consistently outperforms the current state-of-the-art baselines by a margin of 1–3\%.
\end{itemize}

%%%%%%%%%%%%%%%%%%%%%%%%%%%%%%%%%%%%%%%%%%%%%%%%%%%%%%%%%%%%%%%%%%%%%%%%%%%%
% Preliminaries
%%%%%%%%%%%%%%%%%%%%%%%%%%%%%%%%%%%%%%%%%%%%%%%%%%%%%%%%%%%%%%%%%%%%%%%%%%%%
\section{Preliminaries}\label{sec: preliminary}
\paragraph{Hypergraphs.}
A hypergraph is a generalization of a graph in which an edge can join any number of vertices. In contrast, in an ordinary graph, an edge connects exactly two vertices. 
We can denote a hypergraph as $\mathcal{G}=\{{\mathcal{V},\mathcal{E}}\}$, with $|\mathcal{V}|$ nodes and $|\mathcal{E}|$ hyperedges, where $\mathcal{V}=\{v_1, \ldots, v_{|\mathcal{V}|} \}$ is the set of nodes and $\mathcal{E}=\{e_1, \ldots, e_{|\mathcal{E}|} \}$ is the set of hyperedges. 
$\mathcal{E}(i)$ denotes a set containing all the nodes sharing at least one hyperedge with node $i$. 
The incidence matrix $\mathbf{H} \in \mathbb{R}^{|\mathcal{V}| \times |\mathcal{E}|}$ is a common notation in hypergraph. Its entries are defined as $\mathbf{H}_{i,e} = 1$ if node $i$ belongs to hyperedge $e$, and $0$ otherwise.

% We denote the node feature matrix $\mathbf{X} \in \mathbb{R}^{{|\mathcal{V}|} \times d}$ , where the $i$-th row $\mathbf{X}_i$ corresponds to $\mathbf{x}_i$. 

\paragraph{Message Passing in Hypergraphs.}
Neural message passing~\cite{gilmer2017neural,battaglia2018relational} is the most widely used propagator for node feature updates in GNNs, which propagates node features while taking into account their neighboring nodes. Next, we generalize graph message passing to hypergraph, where multiple nodes interaction reflected in a hyperedge is considered, 
\begin{equation}\label{eq:HMP}
    \mathbf{x}_i^{(l+1)} =\Psi^{(l)}\left(\mathbf{x}_i^{(l)},
    \Phi_{2} \left( \{ \mathbf{z}_e^{(l)}, \Phi_{1} (\{\mathbf{x}_j^{(l)}\}_{j\in e})\}_{e\in\mathcal{E}(i)} \right)\right),
\end{equation}
% where $\mathcal{E}(i)$ means the set of hyperedges containing node $i$. 
where $\mathbf{z}_e^{(l)}$ denotes the feature embedding of hyperedge in the $l^{th}$ layer. 
% , and $\mathbf{z}_e^{(0)}$ is typically initialized by a zero vector or the average of the feature vectors of the nodes in this hyperedge. 
$\Phi_1^{(l)}$ and $\Phi_2^{(l)}$ denote differentiable permutation invariant functions (e.g., sum, mean, or max) for nodes and hyperedges, respectively. $\Psi^{(l)}$ denotes a differentiable function such as MLPs.

\paragraph{Neural ODEs.} 
Neural ODEs~\cite{weinan2017proposal, chen2018neural} are essentially ordinary differential equations (ODEs), where the time derivative of the hidden states are parameterized by a neural network $f_{\theta}$, 
\begin{equation}\label{eq:neural_ode}
    \frac{\mathrm{d} \mathbf{x}(t)}{\mathrm{d}t} = f_{\theta} \left(\mathbf{x}{(t)},t\right),\quad \mathbf{x}(0) = \mathbf{x}, 
\end{equation}
where $t$ denotes time, $\mathbf{x}(t)$ are the system state, and $\theta$ is learnable parameter. It can be understood as the continuous limit form of the residual network, e.g, $\mathbf{x}(t+\tau) = f_{\theta} \left(\mathbf{x}{(t)},t\right)+\mathbf{x}(t)$, where $\tau$ denotes the step size of time. Thus, Neural ODEs model the continuous dynamics system that evolve hidden states over a continuous range of ``depths'', analogous to layers in traditional deep networks.

%%%%%%%%%%%%%%%%%%%%%%%%%%%%%%%%%%%%%%%%%%%%%%%%%%%%%%%%%%%%%%%%%%%%%%%%%%%%
% Hypergraph Message Passing based on Particle Dynamics
%%%%%%%%%%%%%%%%%%%%%%%%%%%%%%%%%%%%%%%%%%%%%%%%%%%%%%%%%%%%%%%%%%%%%%%%%%%%
\section{Hypergraph Message Passing based on Particle Dynamics}
Viewing HNNs as continuous systems allows for the application of physical models to elucidate and investigate their properties. A compelling conceptual analogy arises when comparing hypergraph message passing in deep learning with interacting particle systems studied in statistical physics. These disparate fields display remarkable structural and dynamical resemblances: the local potential field-governed interactions among particles in many-body systems are analogous to the aggregation operations in HNNs. Concurrently, the emergent collective behaviors observed in self-organizing dynamics demonstrate a functional equivalence to information propagation mechanisms in HNNs. As message passing delineates the transfer of information between entities, particle systems can represent this by having particles acting as entities that exchange information through the transition of specific attributes. Moreover, such particle systems can be endowed with hypergraph structures by interpreting hyperedges as distinct fields that exert influence over all nodes encompassed by them. 
% We consider HNNs as continuous system and use physical model to explain and study them. An intriguing conceptual parallel emerges between hypergraph message passing in deep learning and interacting particle system studied in statistical physics. They exhibit striking structural and dynamical similarities: particles interaction governed by local potential fields in many-body systems mirrors the aggregation operations of HNNs, while the emergent collective behaviors observed in self-organizing particle dynamics bear functional equivalence to information propagation mechanisms. Since message passing shows how information moves from one entity to another through their connections, we can use particles in a system as entities and let them share information by the transition of some attributes. In addition, the particle system can also have hypergraph structures if we treat hyperedges as different fields that affect all the nodes in certain hyperedges. 

\paragraph{Fields and Hypergraphs.}

The hypergraph uses hyperedges which reflect the common relationship among multiple nodes. Such multi-nodal relationships are aptly described using ``fields''. Each field imposes a common dynamic on the nodes under its influence, which is equivalent to a hyperedge defining a unified information propagation step for its constituent nodes. A node participating in multiple hyperedges thus receives information from diverse sources due to field superposition. Thus, hyperedges reflect both the interactions of particles within a specific region and the concurrent influence of the external environment, which together dictate the propagation dynamics. Based on these observations, we propose a hypergraph message passing framework for describing information dynamics. We formalize a composite field for each node $i$ by $F_i = \sum\limits_{e: i \in e}F_i^e + F_d$ to encapsulate the hypergraph structure. This field $F_i$ aggregates influences akin to those in a particle system, where $F_i^e$ represents the interaction energy for node $i$ with the edge $e$, and $F_d$ is the damping energy.

\paragraph{Interaction Force.}

Interaction force forms a basic association between hypergraph message passing and particle dynamics. We consider particles identified by their specific attributes $\mathbf{p}_i \in \mathbb{R}^{d}$ in the particle system. Then, we define the interaction energy as following 
\begin{equation}\label{eq: model orignal}
\sum\limits_{e:i\in e} F_i^e := \frac{1}{2} \sum\limits_{e:i\in e} \sum\limits_{j\in e}  (\mathbf{p}_i - \mathbf{p}_j)^\top f_\beta(\mathbf{p}_i, \mathbf{p}_j,\mathbf{e}) (\mathbf{p}_i-\mathbf{p}_j), \quad e \in \mathcal{E}.
\end{equation}
If $f_\beta(\mathbf{p}_i,\mathbf{p}_j,e)>0,$ node $i$ is attracted by node $j$. If $f_\beta(\mathbf{p}_i,\mathbf{p}_j,e)<0,$ node $i$ is repulsed by node $j$. They correspond to two basic interaction forces: attraction and repulsion. 
%  ({\color {green} Do you mean in previous model?})
Note that in the previous model similar to \eqref{eq: model orignal}, the only interaction is attraction. 
It relies on an implicit belief that all the hyperedges only connect similar nodes. But this is not always true in the real world. Hence, introducing repulsion into hypergraph neural networks is significant.

\paragraph{Damping Force.}
% In fact, the specific form of $f_d$ can be chosen variously. There are two main reasons to make use of damping in neural networks. One is to keep network system stable. If the repulsive force does not decay quickly enough as the distance between particles increases, it is possible that some particles will separate infinitely. A damping term can avoid network collapse by dominating the dynamics near the boundary. The other reason is that the damping term in many cases can encourage different collective dynamics~\cite{kolokolnikov2011stability,kolokolnikov2013emergent}. 
The specific formulation of the damping term can vary. The inclusion of damping in neural networks serves two principal purposes. Firstly, it is crucial for ensuring system stability. If inter-particle repulsive forces do not attenuate adequately with increasing distance, particles risk unbounded separation or unbounded Dirichlet energy, which could lead to network collapse. A damping term mitigates this by exerting a dominant influence on the dynamics, especially near the system's boundaries. Secondly, the damping term has been demonstrated to promote various collective dynamics in numerous systems~\cite{kolokolnikov2011stability,kolokolnikov2013emergent}. For damping, we take double-well potential $F_d = \zeta(1-\mathbf{p}^2)^2$ with the coefficient $\zeta$. The corresponding damping force is given by the Allen-Cahn force $f_d(\mathbf{p}) = \nabla_v F_d = \delta (1-\mathbf{p}^2)\mathbf{p}$~\cite{rayleigh1894jws,fang2019emergent}, where $\delta$ denotes the strength coefficient. This formulation exhibits favorable separation properties, as will be demonstrated in Section~\ref{os on diffusion}.

\paragraph{Hypergraph Dynamics.}
% Note that the interaction term can be chosen in different methods, they can be roughly divided into two categories: the reductive expression and the non-reductive expression.
% For the reductive expression, convolution coefficients\cite{gao2022hgnn}, attention coefficients\cite{bai2021hypergraph} and so on can be considered. However, the hyperedges reduced to the clique, which will lost high-order information. For the non-reductive expression, we can consider the star expansion, which captures high-order information by connecting each node with the hyperedges it belongs to. 
Interaction term can be formulated using two primary approaches: reductive and non-reductive expressions. Reductive methods, such as those based on convolution coefficients~\cite{gao2022hgnn} or attention coefficients~\cite{bai2021hypergraph}, often simplify hyperedges into clique-like structures, which can lead to the loss of crucial high-order information. In contrast, non-reductive expressions, like star expansion, aim to preserve this high-order information by explicitly connecting each node to the hyperedges it belongs to. We take the non-reductive expression and Allen-Cahn force to design interaction force $f_\beta$ and damping force $f_d$. In fact, the particle attribute $\mathbf{p}$ can represent diverse properties, such as features or velocity. This flexibility in defining $\mathbf{p}$ naturally allows for the formulation of both first-order and second-order particle systems.

\paragraph{First-order ODEs.}
Inspired by the opinions dynamics~\cite{motsch2014heterophilious}, we have $\mathbf{p}_i \mapsto \mathbf{x}_i \in \mathbb{R}^{d}$ that  interact with each other according to the first-order system. Taking gradient of composite field $F$, we have
\begin{equation}\label{eq:1st order}
    \frac{\mathrm{d} \mathbf{x}_i}{\mathrm{d}t} = \sum\limits_{e:i\in e}\sum\limits_{j\in e}  f_\beta(\mathbf{x}_i, \mathbf{x}_j,e) (\mathbf{x}_j - \mathbf{x}_i) + f_d(\mathbf{x}_i),
\end{equation}
where $f_\beta$ is a parameterized function. 
This form has connection with diffusion process of hypergraph. We defer the detailed analysis to Section~\ref{sec:scales}.

\paragraph{Second-order ODEs.} 

Inspired by the flocking dynamics~\cite{motsch2014heterophilious, Linglong_stochastic}, this is second-order model where the attribute is the velocity, $\mathbf{p}_i \mapsto \mathbf{v}_i \in \mathbb{R}^{d}$, which is coupled to the feature $\mathbf{x}_i \in \mathbb{R}^{d}$. In this sense, information can be propagated by the evolution of feature and velocity of nodes, 
\begin{equation}\label{eq:2nd order}
    \begin{aligned}
        \frac{\mathrm{d} \mathbf{v}_i}{\mathrm{d}t} = \sum\limits_{e:i\in e}\sum\limits_{j\in e} f_\beta(\mathbf{x}_i, \mathbf{x}_j,e) (\mathbf{v}_j-\mathbf{v}_i) + f_d(\mathbf{v}_i),  \text{ where } \frac{\mathrm{d} \mathbf{x}_i}{\mathrm{d}t} = \mathbf{v}_i.
    \end{aligned}
\end{equation}

\paragraph{From ODEs to SDEs.} 
To capture incompleteness and uncertainty within hypergraph data, stochasticity is introduced into the particle system. For instance, such uncertainty can arise when particles from different classes exhibit similar features. To model this inherent randomness, these stochastic processes are often described using stochastic differential equations (SDEs). We assume these SDEs are driven by Brownian motion $B_t$: 
\begin{equation}\label{eq: 1st order with noise}
    \mathrm{d} \mathbf{p}(t+\tau) = \nabla F(\mathbf{p}(t)) \mathrm{d}t + \epsilon \mathrm{d} B_t,  \text{ where } \mathbb{E}[\mathrm{d}B_t] = 0 \text{ and } \mathrm{Cov}(\mathrm{d}B_t) = \mathbf{I}_d,  
\end{equation}
% \begin{equation}\label{eq: 2st order with noise}
%     \mathrm{d} \mathbf{v}(t+\tau) = \left[ -\frac{1}{2}\nabla F_\beta(\mathbf{v}(t)) + \nabla F_d(\mathbf{v}(t)) \right] \mathrm{d}t + \epsilon \mathrm{d} B_t, \text{ where } \mathrm{d} \mathbf{x}(t) = \mathbf{v}(t) \mathrm{d}t,
% \end{equation}
% Here, $\epsilon$ is a coefficient for Brownian motion $B_t$, which can be either treated as learnable parameter or hyperparameter, and 
where $\mathbf{I}_d$ denotes the $d \times d$ identity matrix. In essence, SDEs describe the changes in system state under infinitesimal time variations. Here, particle states undergo continuous evolution through both deterministic drift term and stochastic diffusion component. It can describe hypergraph evolution of neural message passing in more realistic scenario. 
According to~\cite{Linglong_stochastic, ha2009emergence}, this model can achieve self-organization in a finite time, which enables the system to adjust its own state in the shortest time.

\paragraph{Hypergraph Message Passing.} 

We introduce a unified hypergraph message passing framework, called \emph{Hypergraph Atomic Message Passing}, or \textsf{HAMP}, given  in \eqref{eq: 1st order with noise}. By instantiating this framework with either a first‐order or a second‐order system formulation, we obtain two variants: \textsf{HAMP-I} and \textsf{HAMP-II}, respectively. Detailed implementations of both approaches are provided systematically in Appendix \ref{alg} to ensure clarity and maintain manuscript focus.

%%%%%%%%%%%%%%%%%%%%%%%%%%%%%%%%%%%%%%%%%%%%%%%%%%%%%%%%%%%%%%%%%%%%%%%%%%%%
% Scale translation of hypergraph: diffusion and particle dynamics
%%%%%%%%%%%%%%%%%%%%%%%%%%%%%%%%%%%%%%%%%%%%%%%%%%%%%%%%%%%%%%%%%%%%%%%%%%%%
\section{Scale Translation of Hypergraph: Diffusion and Particle Dynamics}\label{sec:scales}

We examine hypergraph message passing through two complementary lenses—microscopic and macroscopic. From the microscopic viewpoint, \textsf{HAMP} is cast as a particle‐dynamics system, as previously discussed. From the macroscopic standpoint, message propagation is seen as a diffusion process, which reveals that the particle‐based formulation naturally subsumes diffusion‐based models and overcomes their intrinsic limitations.

\paragraph{Hypergraph Diffusion.}

Consider node feature space $\Omega=\mathbb{R}^d$ and tangent vector field space $T\Omega=\mathbb{R}^d$. For $\mathbf{x,y}\in\Omega$, $\mathfrak{x,y}\in T\Omega$, and $\mathfrak{x}_{i,j}=-\mathfrak{x}_{j,i}$, we adopt the following inner products:
\begin{equation}
\left<\mathbf{x},\mathbf{y}\right>=\sum\limits_{i,j}\mathbf{x}_i\mathbf{y}_j, \quad [\mathfrak{x},\mathfrak{y}]=\sum\limits_{i>j}\sum\limits_{e\in\mathcal{E}} h_{i,j}^e\:\mathfrak{x}_{i,j}\mathfrak{y}_{i,j}.
\end{equation}
Here $h_{i,j}^e$ is a tuple related to node $i,j$ and hyperedge $e$ and $h_{i,j}^e=0$ if $\mathbf{H}_{i,e}\mathbf{H}_{j,e}=0.$ We set $h_{i,j}^e$ to satisfy $\sum\limits_{j}\sum\limits_{e\in\mathcal{E}}h_{i,j}^e=1$. For any $\mathfrak{u}\in T\Omega,$ by the adjoint relation $[\mathfrak{u},\nabla\mathbf{x}] = \left<\mathbf{x},\text{div}\,\mathfrak{u}\right>$, 
where $\nabla \mathbf{x} = \mathbf{x}_j -\mathbf{x}_i,$ we can derive
$(\text{div}\mathfrak{u})_j = \sum\limits_{i} \sum\limits_{e\in\mathcal{E}}h_{i,j}^e u_{i}$. 
Then, the hypergraph diffusion process is %given by
\begin{equation}\label{eq: diffusion}
    \frac{\mathrm{d}\mathbf{x}_i}{\mathrm{d}t} = \text{div}\nabla\mathbf{x}_i = \sum\limits_{j} \sum\limits_{e\in\mathcal{E}}h_{i,j}^e (\mathbf{x}_j-\mathbf{x}_i).
\end{equation}
For simplicity, we rewrite \eqref{eq: diffusion} in matrix form $\frac{d\mathbf{x}}{\mathrm{d}t} =-\mathcal{L}\mathbf{x}$,
where $\mathcal{L}=\mathbf{I}-(\sum\limits_{e\in\mathcal{E}}h_{i,j}^e)$ is a hypergraph operator. If $\mathcal{L}$ is positive semi-definite, we interpret \eqref{eq: diffusion} as a diffusion-type process of hypergraph. Different parameterizations of $h_{i,j}^e$ then yield distinct diffusion equations. For example, applying a forward Euler discretization to \eqref{eq: diffusion} and setting $(\sum\limits_{e\in\mathcal{E}}h_{i,j}^e)= \mathbf{D}_v^{-\frac{1}{2}}\mathbf{H} \mathbf{W} \mathbf{D}_e^{-1}\mathbf{H}^{\top} \mathbf{D}_v^{-\frac{1}{2}}$ recovers the simplified HGNN~\cite{feng2019hypergraph} without channel mixing.

\paragraph{Connecting Particle Dynamics.}

Since \eqref{eq: diffusion} coincides with the self-organized dynamics in particle system~\cite{motsch2014heterophilious}, we reinterpret \eqref{eq: diffusion} not as a standard hypergraph diffusion process, but rather as particle dynamics where $h_{i,j}^e$ denotes the interaction force between nodes $i$ and $j$ under field $e$. In fact, \eqref{eq: diffusion} is a special case of \eqref{eq:1st order} where only attractive forces are considered in the message propagation. As we show in Section~\ref{os on diffusion}, this simplification is atypical in particle systems and leads to over-smoothing.
% —a simplification that, as we show in , is atypical in particle systems and leads to over-smoothing.

%%%%%%%%%%%%%%%%%%%%%%%%%%%%%%%%%%%%%%%%%%%%%%%%%%%%%%%%%%%%%%%%%%%%%%%%%%%%
% Theory of Anti-over-smoothing
%%%%%%%%%%%%%%%%%%%%%%%%%%%%%%%%%%%%%%%%%%%%%%%%%%%%%%%%%%%%%%%%%%%%%%%%%%%%
\section{Theory of Anti-over-smoothing}\label{os on diffusion}

For diffusion-type hypergraph networks, we define the \emph{hypergraph Dirichlet energy} of $\mathbf{x}\in\mathbb{R}^{N\times d}$ as 
% \begin{equation}
$\mathbf{E}(\mathbf{x}) := \sum\limits_{i,j=1}^{N} \sum\limits_{e\in \mathcal{E}} \mathbf{H}_{i,e} \mathbf{H}_{j,e}\|\mathbf{x}_i-\mathbf{x}_j\|^2 = \textrm{tr}(\mathbf{x}^\top\mathcal{L}\mathbf{x})$.
% \end{equation}
% \begin{remark}
%     Actually, one can define hypergraph Dirichlet energy by $\mathbf{E}(\mathbf{x}):= \textrm{tr}(\mathbf{x}^\top\mathcal{L}\mathbf{x})$ related to the Laplacian $\mathcal{L}.$ 
%     % However, since $\mathcal{L}$ is not a deterministic matrix here, we choose a simpler definition like~\cite{rusch2022graph}. The key point is that Dirichlet energy can describe the difference among all the node features. Hence the simplification is acceptable.
%     % For vector field $\mathbf{x}=[\mathbf{x}_1,\cdots,\mathbf{x}_N],$ Dirichlet energy is defined as 
%     % \begin{equation}
%     %     \mathbf{E}(\mathbf{x}):= \textrm{tr}(\mathbf{x}^\top\mathcal{L}\mathbf{x})
%     % \end{equation}
% \end{remark}
Furthermore, we will define over-smoothing in hypergraph message passing.
\begin{definition}
    Let $\mathbf{x}^{(l)}$ denote the hidden features of the $l^{th}$ layer. We define over-smoothing in HNNs as the exponential convergence to zero of the layer-wise Dirichlet energy as a function of $l$, i.e., $\mathbf{E}(\mathbf{x}^{(l)})\le C_1 e^{-C_2 l}$, with positive constants $C_1$ and $C_2$.
\end{definition}

\paragraph{Why Do HGNN Cause Over-smoothing?}

The normalized hypergraph Laplacian matrix is defined by $\mathcal{L} = \mathbf{I} - \mathbf{P}$, where $\mathbf{P}$ is the propagation matrix derived from the incidence matrix $\mathbf{H}$. In an HGNN, the feature at layer $l$ without activation  $\sigma$ evolves according to $\mathbf{x}^{(l)} = \mathbf{P}^{l-1}\mathbf{x}^{(0)} \Theta^{(1)} \cdots \Theta^{(l-1)}$. However, this purely diffusive propagation inevitably induces over-smoothing: one can show $\mathbf{E}(\mathbf{x}^{(l)}) \le C e^{-\gamma^2 l}$~\cite{chamberlain2021grand}, 
% ({\color {green} cite where this is shown})
where $\gamma$ is the smallest non-zero positive eigenvalue of $\mathcal{L}.$ This over-smoothing behavior stems from the intrinsic diffusion dynamics. Because $\mathcal{L}$ is symmetric positive semi-definite, repeated application of $\mathbf{P}$ causes node representations to decay exponentially towards a limiting state--which is zero, thereby eroding discriminative power.
  
% It is the diffusion structure that give rise to over-smoothing, because the representation of node feature $\mathbf{x}^{(l)}$ cannot avoid decaying to zero exponentially under a symmetric positive semi-definite kernel $\mathcal{L}$.  Compared to HGNN, ED-HNN uses the random walk normalized hypergraph Laplacian matrix and incorporates additional MLP modules. Furthermore, the initial residual connection is integrated to explicitly counteract some of feature over-smoothing. Existing HNNs variants, such as Deep-HGNN~\cite{chen2022preventing} and TF-HNN~\cite{tang2025trainingfree}, adopt similar graph-based methodologies to mitigate the issue of over-smoothing.

%, where the symmetrically normalized propagation matrix is defined as $\mathbf{P} = \mathbf{D}_v^{-\frac{1}{2}}\mathbf{H} \mathbf{W} \mathbf{D}_e^{-1}\mathbf{H}^{\top} \mathbf{D}_v^{-\frac{1}{2}}$. 
% In other words, the node representations $\mathbf{x}^{(l)}$ in deep layer converge to a distribution that is independent of the input node features. 
%~\cite{cai2020note} show that dropping edges and randomly increasing the weight of a few edges to a high value will help to relieve over-smoothing.

\paragraph{Why Do \textsf{HAMP} Avoid Over-smoothing?}

We now derive theoretical guarantees for the collective behavior of models \eqref{eq:1st order} and \eqref{eq:2nd order}. They show that the addition of a repulsive force enforces a positive lower bound on the Dirichlet energy. Technically, we suppose there exists $\{f_\beta^e\}$ such that $\mathcal{I}=\{1,\cdots,N\}$ can be divided into two disjoint groups with $N_1,N_2$ particles respectively: 
$f_{\beta}(h_{i,j}^e)\ge 0,$ for $\{i,j\}\in \mathcal{I}_1$ or $\mathcal{I}_2$ and $ f_{\beta}(h_{i,j}^e)\le 0,$ otherwise. 
We designate $\{x^{(1)}_i\}:=\{\mathbf{x}_i | i\in\mathcal{I}_1\}$ and $\{x^{(2)}_j\}:=\{\mathbf{x}_j | j\in\mathcal{I}_2\}$.  
Finally, we impose the symmetry $f_{\beta}(h^e_{i,j})=f_\beta(h^e_{j,i})$, which reflects equal-and-opposite interactions under the same field. Under these conditions, one can show that the Dirichlet energy of our system admits a strictly positive lower bound, as follows. We leave the detailed proofs in Appendix~\ref{proof}.

\begin{proposition}[$L_2$ separation of \textsf{HAMP-I}]\label{thm:L2 separation}
For \eqref{eq:1st order}, suppose the above assumptions are satisfied.  Define the mean value $\mathbf{\Bar{x}} := \frac{1}{N}\sum\limits_{i=1}^N \mathbf{x}_i$, 
% the deviation value $\mathbf{\hat{x}}_i := \mathbf{x}_i - \mathbf{\Bar{x}}$,
and the second moments $M_2(\mathbf{x}):= \sum \limits_{i=1}^N \mathbf{x}_i^2.$ Then for sufficiently large $N_1,N_2$, there exist constants $\lambda_- , \lambda_+, $ such that if the initial data satisfies 
% \begin{equation}
    $\lambda(0) := \frac{\widehat{M}_2(0)}{\|\mathbf{\Bar{x}}^{(1)}(0)-\mathbf{\Bar{x}}^{(2)}(0)\|^2} \le \lambda_+$,
% \end{equation}
then, there holds that  the $L_2$  separation
\begin{equation}
    \lambda(t):= \frac{\widehat{M}_2(t)}{\|\mathbf{\Bar{x}}^{(1)}(t) - \mathbf{\Bar{x}}^{(2)}(t)\|^2}
    \le \lambda_- + (\lambda(0)-\lambda_-)e^{-\mu t},
\end{equation}
with a positive constant $\mu$, where $\widehat{M}_2(t):=M_2(\mathbf{x}^{(1)}(t))+M_2(\mathbf{x}^{(2)}(t))$.
\end{proposition}

\begin{proposition}[$L_2$ separation of \textsf{HAMP-II},~\cite{fang2019emergent}]\label{thm:L2 separation of 2nd}
    For \eqref{eq:2nd order}, we set $0<S\le f_{\beta}(h_{i,j}^e)$ for $\{i,j\}\in \mathcal{I}_1$ and $0\le f_{\beta}(h_{i,j}^e)\le D$ otherwise, with $k:= \max\limits_i\{|\mathcal{E}(i)|\}$. If the initial $\|\mathbf{\Bar{x}}^{(1)}(0)-\mathbf{\Bar{x}}^{(2)}(0)\|\gg1,$ and if there exists a positive constant $\eta$ such that
    \begin{equation}\label{eq:biflocking-cond}
          \alpha (S-D)k \min\{N_1,N_2\}\geq \delta+\eta,
    \end{equation}
    Then the system has a bi-flocking.
\end{proposition}

\begin{proposition}[Lower bound of the Dirichlet energy]\label{prop: lower bound}
     If the hypergraph $\mathcal{H}$ is a connected one, for \eqref{eq:1st order} with the conditions of Theorem \ref{thm:L2 separation}, or for \eqref{eq:2nd order} with conditions of Theorem 5.1 in~\cite{fang2019emergent}, there exists a positive lower bound of the Dirichlet energy.
\end{proposition}

%%%%%%%%%%%%%%%%%%%%%%%%%%%%%%%%%%%%%%%%%%%%%%%%%%%%%%%%%%%%%%%%%%%%%%%%%%%%
% Experiment
%%%%%%%%%%%%%%%%%%%%%%%%%%%%%%%%%%%%%%%%%%%%%%%%%%%%%%%%%%%%%%%%%%%%%%%%%%%%
\section{Experiments}\label{Exp}

\subsection{Experiment Setup}
We conduct comprehensive experiments to evaluate the proposed models on node classification task. For more experimental details such as datasets and hyperparameters, please refer to Appendix~\ref{ex_details}.

\paragraph{Datasets.} 
Following ED-HNN\cite{wang2022equivariant}, the real-world hypergraph benchmarking datasets span diverse domains, scales, and heterophiilic levels. They can be divided into two groups based on homophily. The homophilic hypergraphs include academic citation networks (Cora, Citeseer, and Pubmed) and co-authorship networks (Cora-CA and DBLP-CA). 
%For example, Pubmed exhibits the highest CE homophily, indicating strong class consistency within hyperedges. 
The heterophilic hypergraphs cover legislative voting records (Congress, House, and Senate) and retail relationships (Walmart).

\paragraph{Baselines.}
The selected baselines cover two types of hypergraph learning frameworks, comprising both reductive and non-reductive approaches. The reductive methods include HGNN~\cite{feng2019hypergraph}, HCHA~\cite{bai2021hypergraph}, HNHN~\cite{dong2020hnhn}, HyperGCN~\cite{yadati2019hypergcn}, and HyperND~\cite{prokopchik2022nonlinear}. The non-reductive methods include UniGCNII~\cite{ijcai21UniGNN}, AllDeepSets~\cite{chien2022you}, AllSetTransformer~\cite{chien2022you}, ED-HNN~\cite{wang2022equivariant}, and HDS$^{ode}$~\cite{yan2024hypergraph}.

\subsection{Node Classifications on Hypergraphs}
In this section, we evaluate \textsf{HAMP-I} and \textsf{HAMP-II} on nine real-world hypergraph benchmarks for the node classification task. \tabref{tab:NC_exp} reports the accuracy on both homophilic and heterophilic datasets. As HDS$^{ode}$ does not provide results on these benchmarks, we reproduce its performance using the official open-source code and perform hyperparameter tuning following the original paper. For other baselines, our results are consistent with those reported by ED-HNN. Overall, our models demonstrate competitive performance across all nine datasets. Notably, the improvement is more pronounced on heterophilic datasets, with the largest accuracy gain of 3\% observed on Walmart dataset. 

\begin{table*}[!]
    \caption{Node Classification on standard hypergraph benchmarks. The accuracy (\%) is reported with a standard deviation from 10 repetitive runs. (Key: \textbf{Best}; \underline{Second Best}; \dashuline{Third Best.})}
    \label{tab:NC_exp}
    \begin{center}
    \resizebox{0.92\textwidth}{!}{
    \begin{tabular}{lccccc}
    \toprule
    Homophilic  &Cora  &Citeseer   &Pubmed  &Cora-CA &DBLP-CA  \\
    \midrule
    HGNN        
    &79.39$\pm$1.36 &72.45$\pm$1.16 &86.44$\pm$0.44 &82.64$\pm$1.65 &91.03$\pm$0.20  \\
    HCHA        
    &79.14$\pm$1.02 &72.42$\pm$1.42 &86.41$\pm$0.36 &82.55$\pm$0.97 &90.92$\pm$0.22  \\
    HNHN        
    &76.36$\pm$1.92 &72.64$\pm$1.57 &86.90$\pm$0.30 &77.19$\pm$1.49 &86.78$\pm$0.29  \\
    HyperGCN    
    &78.45$\pm$1.26 &71.28$\pm$0.82 &82.84$\pm$8.67 &79.48$\pm$2.08 &89.38$\pm$0.25  \\
    UniGCNII    
    &78.81$\pm$1.05 &73.05$\pm$2.21 &88.25$\pm$0.40 &83.60$\pm$1.14  &\underline{91.69$\pm$0.19} \\
    HyperND     
    &79.20$\pm$1.14 &72.62$\pm$1.49 &86.68$\pm$0.43 &80.62$\pm$1.32 &90.35$\pm$0.26  \\
    AllDeepSets 
    &76.88$\pm$1.80 &70.83$\pm$1.63 &88.75$\pm$0.33 &81.97$\pm$1.50 &91.27$\pm$0.27  \\
    AllSetTransformer  
    &78.58$\pm$1.47 &73.08$\pm$1.20 &88.72$\pm$0.37 &83.63$\pm$1.47 &91.53$\pm$0.23  \\
    ED-HNN     
    &80.31$\pm$1.35 &73.70$\pm$1.38 &\underline{89.03$\pm$0.53} &83.97$\pm$1.55  &\textbf{91.90$\pm$0.19}    \\
    HDS$^{ode}$ 
    &\dashuline{80.65$\pm$1.22} &\dashuline{74.87$\pm$1.12} &88.81$\pm$0.43 &\underline{84.95$\pm$0.98} &91.49$\pm$0.25  \\
    \midrule
    \textsf{HAMP-I}
    &\textbf{81.18$\pm$1.30}   &\underline{75.22$\pm$1.62}  &\dashuline{89.02$\pm$0.38}   &\textbf{85.23$\pm$1.15}   &91.66$\pm$0.17   \\
    \textsf{HAMP-II}
    &\underline{80.80$\pm$1.62}  &\textbf{75.33$\pm$1.61} &\textbf{89.05$\pm$0.41} &\dashuline{84.89$\pm$1.53}  &\dashuline{91.67$\pm$0.23}   \\
    \bottomrule
    \end{tabular}
    }
    \end{center}

    \begin{center}
    \resizebox{0.8\textwidth}{!}{
    \small
    \begin{tabular}{lcccc}
    \toprule
    Heterophilic  &Congress &Senate &Walmart &House \\
    \midrule
    HGNN 
    &91.26$\pm$1.15 &48.59$\pm$4.52 &62.00$\pm$0.24 &61.39$\pm$2.96   \\
    HCHA 
    &90.43$\pm$1.20 &48.62$\pm$4.41 &62.35$\pm$0.26 &61.36$\pm$2.53   \\
    HNHN 
    &53.35$\pm$1.45 &50.93$\pm$6.33 &47.18$\pm$0.35 &67.80$\pm$2.59   \\
    HyperGCN 
    &55.12$\pm$1.96 &42.45$\pm$3.67 &44.74$\pm$2.81 &48.32$\pm$2.93   \\
    UniGCNII 
    &94.81$\pm$0.81 &49.30$\pm$4.25 &54.45$\pm$0.37 &67.25$\pm$2.57   \\
    HyperND 
    &74.63$\pm$3.62 &52.82$\pm$3.20 &38.10$\pm$3.86 &51.70$\pm$3.37    \\
    AllDeepSets  
    &91.80$\pm$1.53 &48.17$\pm$5.67 &64.55$\pm$0.33 &67.82$\pm$2.40   \\
    AllSetTransformer
    &92.16$\pm$1.05 &51.83$\pm$5.22 &65.46$\pm$0.25 &69.33$\pm$2.20    \\
    ED-HNN 
    &\dashuline{95.00$\pm$0.99} &64.79$\pm$5.14 &\dashuline{66.91$\pm$0.41} &\dashuline{72.45$\pm$2.28 }   \\
    HDS$^{ode}$ 
    &90.91$\pm$1.52 &\dashuline{66.90$\pm$5.52} &63.38$\pm$0.48 &71.30$\pm$1.90  \\
    \midrule
    \textsf{HAMP-I}
    &\underline{95.09$\pm$0.79}   &\underline{69.44$\pm$6.09}   &\underline{69.90$\pm$0.38}   &\textbf{72.72$\pm$1.77} \\
    \textsf{HAMP-II}
    &\textbf{95.26$\pm$1.34}  &\textbf{70.14$\pm$6.08} &\textbf{69.94$\pm$0.37} &\underline{72.60$\pm$1.23}    \\
    \bottomrule
    \end{tabular}
    }
    \end{center}
\end{table*}

\subsection{Ablation Studies}\label{ab_studies}
In this section, we conduct several ablation experiments on real-world datasets to assess our model design, and provide empirical validation for our theoretical findings. For more ablation experiments, see Appendix~\ref{ab_details}.

\begin{figure}[!htbp]
\centering
    \subcaptionbox*{} {
    \includegraphics[width=0.32\textwidth]
    {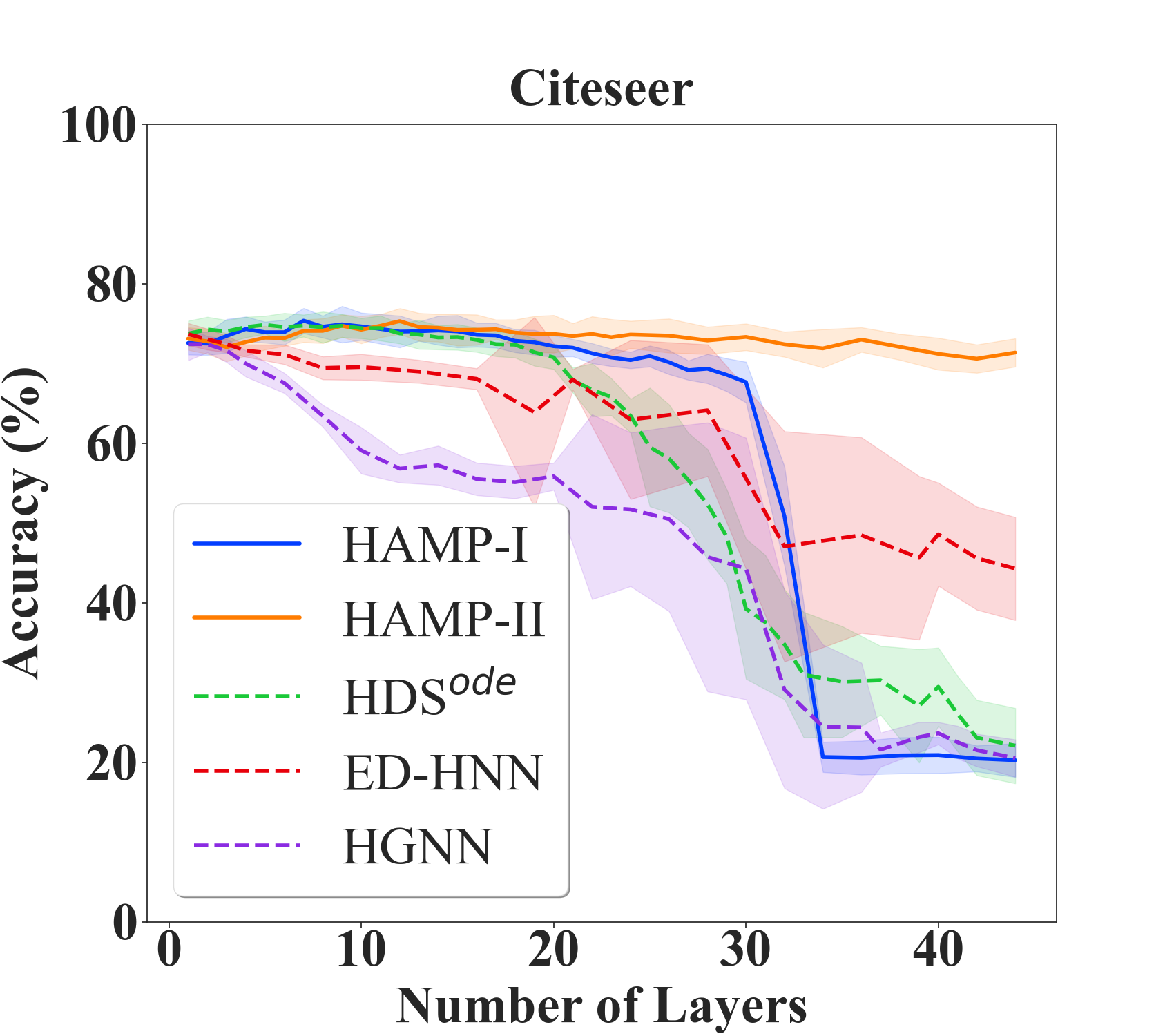}}
    \subcaptionbox*{} {
    \includegraphics[width=0.32\textwidth]
    {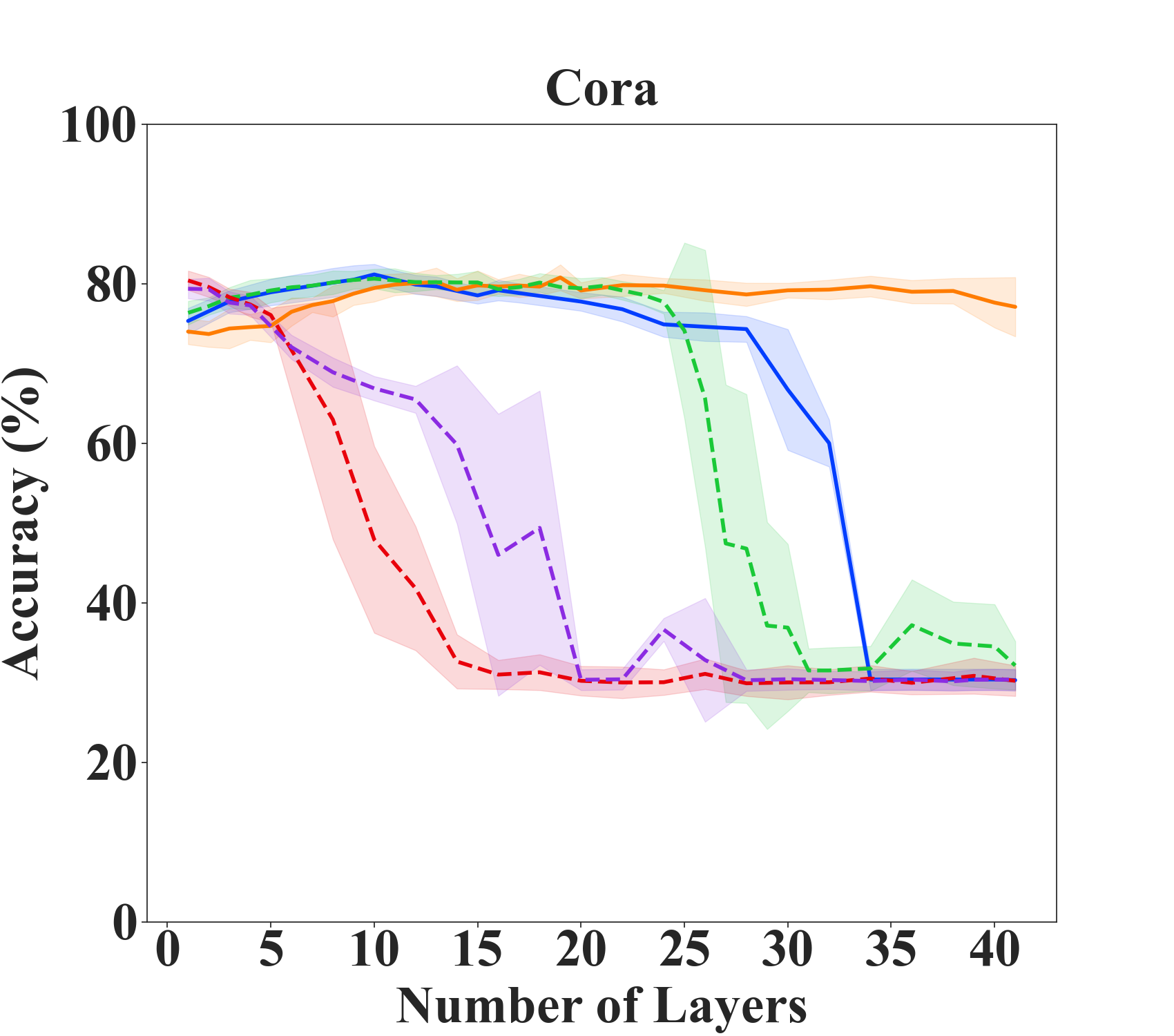}}
    \subcaptionbox*{} {
    \includegraphics[width=0.32\textwidth]
    {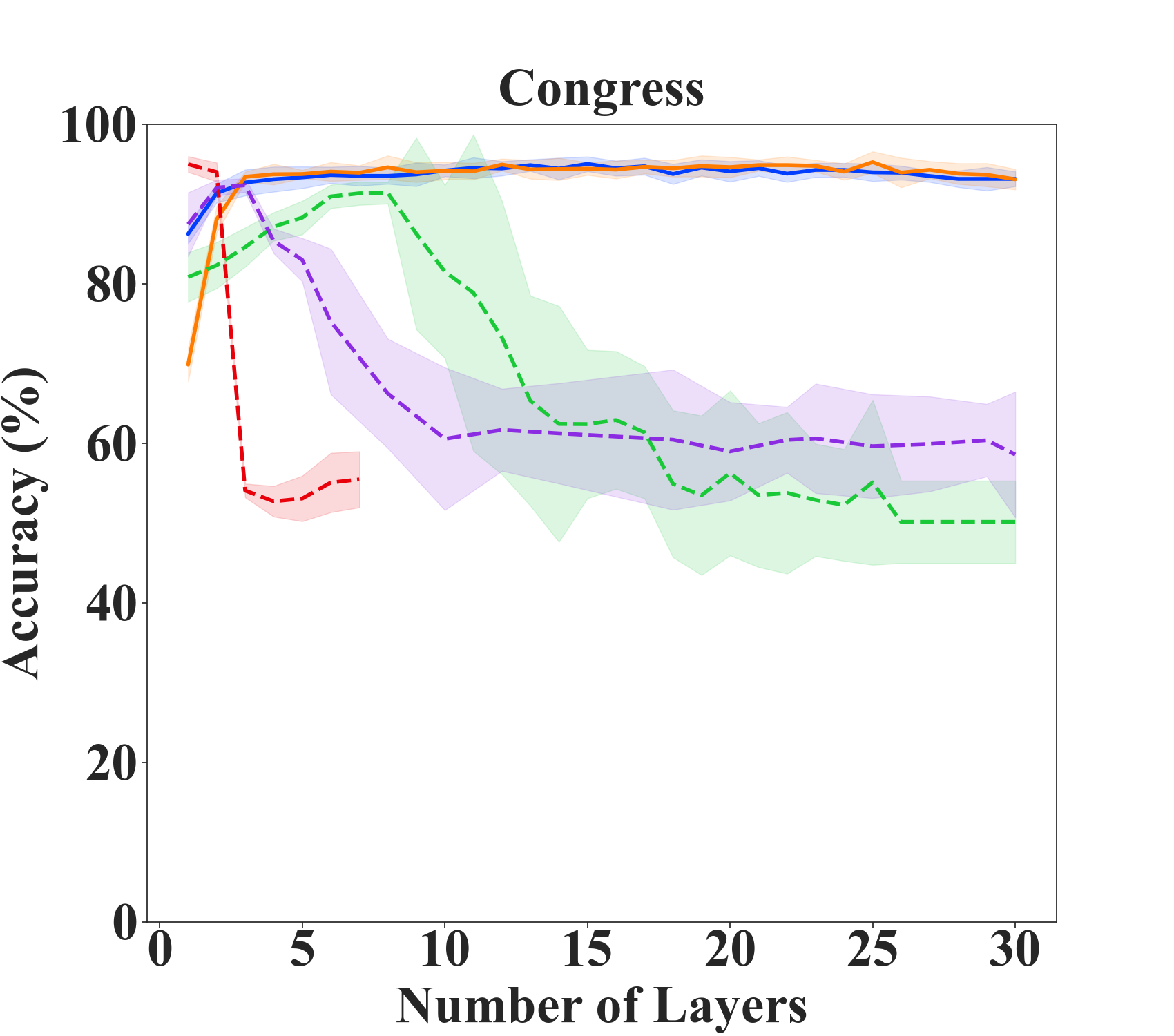}}\vspace{-5mm}
    \caption{An empirical analysis of the depth-accuracy correlation in deep neural networks. The shaded area represents the standard deviation, helping to show the range of accuracy fluctuations.
    }\label{graph:3}
\end{figure}

\paragraph{Impact of the Number of Layers on \textsf{HAMP-I} and \textsf{HAMP-II}.}\label{ex:layers}
To assess the effectiveness of different methods in deep HNNs, we compare three representative baselines with our proposed \textsf{HAMP} models. Unlike Pubmed, DBLP-CA, Senate, and House, the Cora, Citeseer, and Congress datasets generally perform better in shallow networks, but worse in deep networks. Therefore, we focus on these datasets to demonstrate \textsf{HAMP}’s advantages in deep architectures. As shown in \figref{graph:3}, \textsf{HAMP-II} consistently outperforms other methods as depth increases, while competitors suffer from accuracy drops. This ability highlights the potential of \textsf{HAMP} to capture complex representations and maintain stability, making it a valuable framework for deep HNNs. 
% This demonstrates \textsf{HAMP}’s strength in learning complex representations and maintaining stability in deep networks. 
% In summary, our findings suggest that deep learning model, when effectively designed and optimized as demonstrated by \textsf{HAMP}, can significantly enhance the quality of representation learning in challenging datasets.

\paragraph{Impact of the  Repulsion and Allen-Cahn Forces on \textsf{HAMP-I} and \textsf{HAMP-II}.}
In addition to the theoretical analysis, we conducted ablation studies to investigate the individual and combined effects of the repulsion force  $f_\beta^-$ and the  Allen-Cahn force $f_d$ on both \textsf{HAMP-I} and \textsf{HAMP-II}. The results given in \tabref{tab:EX_Ablation} show that incorporating the repulsive force significantly improves classification performance. In \textsf{HAMP-I}, enabling repulsion alone yields notable gains over the baseline, while the Allen-Cahn term  alone offers moderate improvements. Notably, combining both terms consistently achieves the best accuracy across all datasets. The synergy between the repulsion and Allen-Cahn terms further boosts performance, confirming that these particle system-inspired mechanisms play complementary roles: repulsion term  prevents feature over-smoothing by separating node embeddings, whereas Allen-Cahn term balances attraction and repulsion to promote class-dependent equilibrium.
% It can be seen that the repulsive force has an obvious effect on improving the node classification performance. In \textsf{HAMP-II}, model with repulsive forces consistently outperformed the other models. These findings confirm the validity of the \textsf{HAMP} construction and further highlight the significant advantages of incorporating particle system theory into the hypergraph message passing learning process.

\begin{table*}[t]
    \small
    \caption{Node Classification on some standard hypergraph benchmarks. The accuracy (\%) is reported from 10 repetitive runs. (Key: $f_\beta^-$: repulsion; $f_d$: Allen-Cahn; \textbf{Best}.)}
    \label{tab:EX_Ablation}
    \begin{center}
    \begin{tabular}{lcc|ccccccc}
    \toprule
      &$f_\beta^-$  &$f_d$   &Cora  &Citeseer &Pubmed &Congress &Senate &Walmart &House \\
    \midrule
    \multirow{4}*{\textsf{HAMP-I}}   
    &\faTimes &\faTimes  &75.67 &70.59 &87.93 &93.47 &60.14 &69.86 &69.88 \\
    &\faCheck &\faTimes  &75.97 &70.60 &88.23 &93.65 &61.69 &69.73 &69.57 \\
    &\faTimes &\faCheck  &80.59 &74.67 &88.77 &94.67 &65.63 &69.73 &71.55  \\
    &\faCheck &\faCheck  &\textbf{81.18}  &\textbf{75.22}  &\textbf{89.02} &\textbf{95.09}   &\textbf{69.44}  &\textbf{69.90}  &\textbf{72.72} \\
    \midrule
    \multirow{4}*{\textsf{HAMP-II}}  
    &\faTimes &\faTimes &77.18 &71.75 &88.68  &94.35 &60.14 &69.84  &70.46 \\
    &\faCheck &\faTimes &77.40 &71.69 &88.77  &94.63 &59.58 &69.80  &69.63 \\
    &\faTimes &\faCheck &79.50 &74.25 &88.80  &94.12 &64.51 &69.86  &70.96 \\
    &\faCheck &\faCheck &\textbf{80.80}   &\textbf{75.33}  &\textbf{89.05}    &\textbf{95.26}  &\textbf{70.14}  &\textbf{69.94} &\textbf{72.60} \\
    \bottomrule
    \end{tabular}
    \end{center}\vspace{-5mm}
\end{table*}

\begin{wrapfigure}[14]{r}{0.33\textwidth} 
\raggedleft
\includegraphics[width=0.33\textwidth]{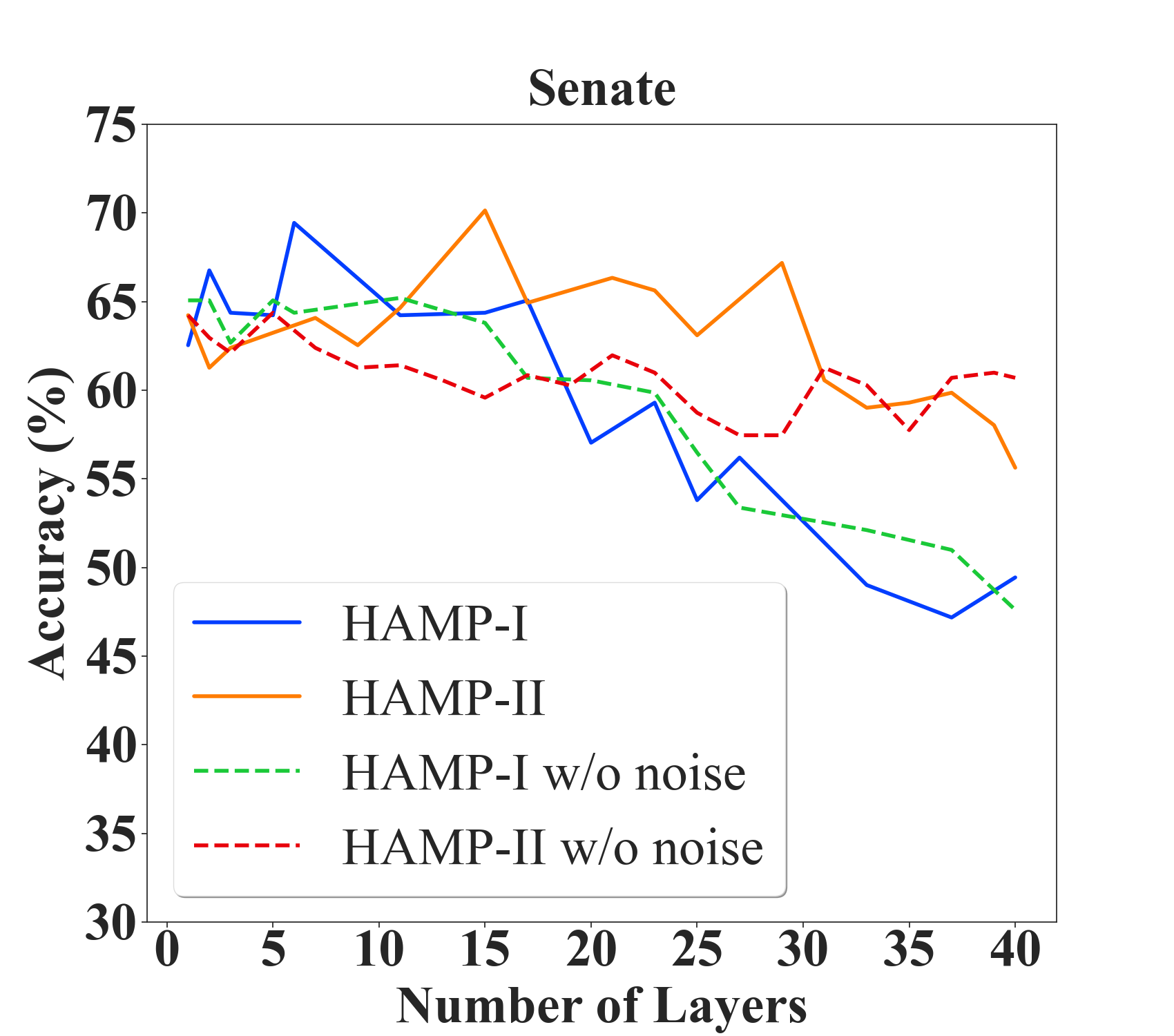}
\caption{Significance plot for noise on Senate dataset.}
\label{fig:noise}
\end{wrapfigure}
\paragraph{Impact of Noise on \textsf{HAMP-I} and \textsf{HAMP-II}.}
\figref{fig:noise} illustrates the effect of adding stochastic component into deterministic message passing on Senate dataset. Notably, incorporating noise consistently improves accuracy for both \textsf{HAMP-I} and \textsf{HAMP-II}. In addition, the stability of \textsf{HAMP-II} is better than that of \textsf{HAMP-I}, whether or not noise is injected, indicating that the second-order particle system is more stable. Overall, these results demonstrate that incorporating stochastic component into message passing effectively improves model performance and stability by explicitly capturing data uncertainty.

\subsection{Vertex Representation Visualization}
\begin{figure}[!htbp]
\centering
    \subcaptionbox*{(a) At epoch = 1.} {
    \includegraphics[width=0.18\textwidth]{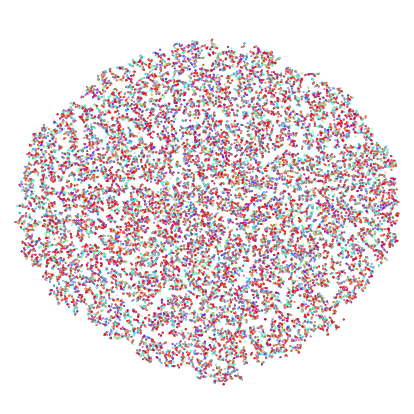}}
    \subcaptionbox*{(b) At epoch = $\frac{1}{8}E$.} {
    \includegraphics[width=0.18\textwidth]{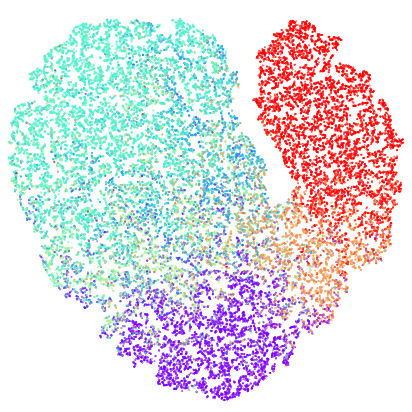}}
    \subcaptionbox*{(c) At epoch = $\frac{1}{4}E$.} {
    \includegraphics[width=0.18\textwidth]{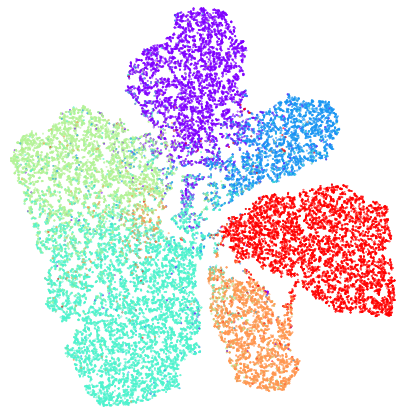}}
    \subcaptionbox*{(d) At epoch = $\frac{1}{2}E$.} {
    \includegraphics[width=0.18\textwidth]{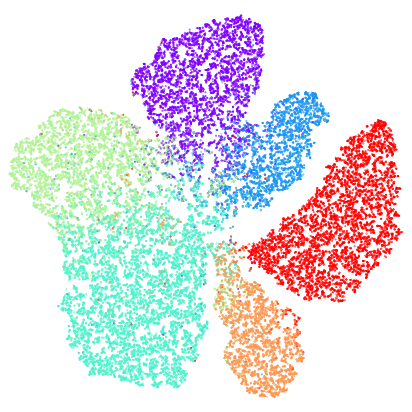}}
    \subcaptionbox*{(e) At epoch = $E$.} {
    \includegraphics[width=0.18\textwidth]{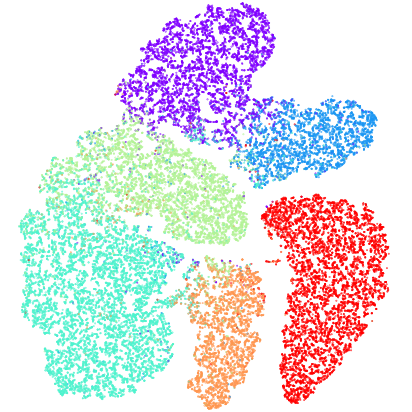}}
    \subcaptionbox*{(f) At epoch = 1.} {
    \includegraphics[width=0.18\textwidth]{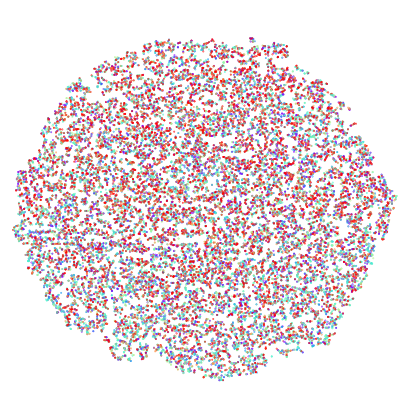}}
    \subcaptionbox*{(g) At epoch = $\frac{1}{8}E$.} {
    \includegraphics[width=0.18\textwidth]{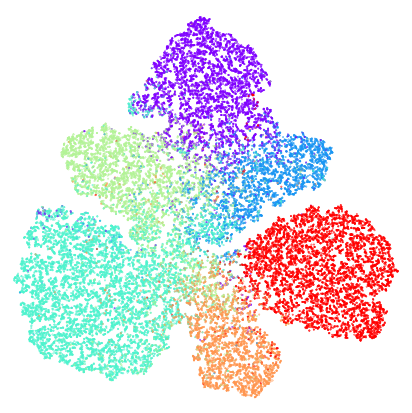}}
    \subcaptionbox*{(h) At epoch = $\frac{1}{4}E$.} {
    \includegraphics[width=0.18\textwidth]{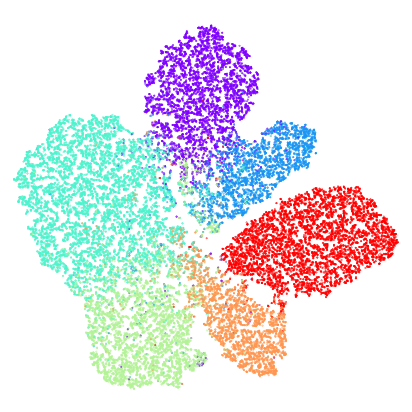}}
    \subcaptionbox*{(i) At epoch = $\frac{1}{2}E$.} {
    \includegraphics[width=0.18\textwidth]{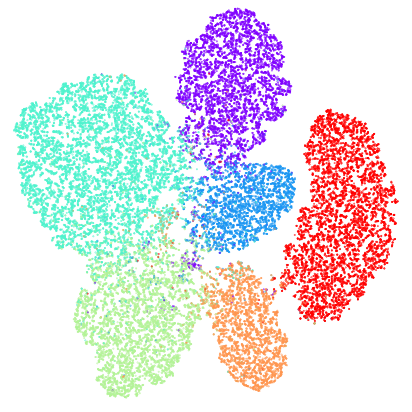}}
    \subcaptionbox*{(j) At epoch = $E$.} {
    \includegraphics[width=0.18\textwidth]{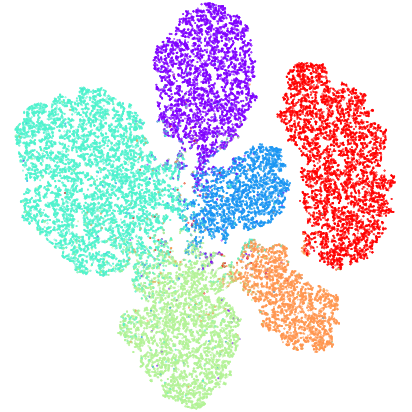}}
    \caption{The t-SNE visualization of vertex representation evolution of \textsf{HAMP-I} (the first row) and \textsf{HAMP-II} (the second row) on Congress dataset. 
    The colors represent the class labels.
    } 
    \label{graph:4}
\end{figure}

To more intuitively validate the progressive refinement of vertex representations in our \textsf{HAMP} methods, we use t-SNE~\cite{vandermaaten08a} to visualize the vertex evolution process of \textsf{HAMP-I} and \textsf{HAMP-II} on Congress dataset at different epochs. As shown in \figref{graph:4}, we visualize the vertices based on the representations obtained at epoch $1$, $\frac{1}{8}E$, $\frac{1}{4}E$, $\frac{1}{2}E$, and $E$, where $E$ is the total number of epochs. From \figref{graph:4}, we have the following three observations:
 \begin{itemize}
     \item When epoch = $1$ (subgraphs (a) and (f)), the node feature representations exhibit a chaotic distribution, and it is difficult to distinguish the number of categories. As training progresses, the clustering entropy shows a monotonically decreasing trend. 
     \item Subgraphs (e) and (j) show the visualization results at convergence for \textsf{HAMP-I} and \textsf{HAMP-II}, respectively. The final category boundaries are clearer in \textsf{HAMP-II} than in \textsf{HAMP-I}, reflecting the geometry refinement enabled by \textsf{HAMP-II}’s deeper message passing.
     \item Comparing subgraphs (b) and (g), \textsf{HAMP-I} is still in the early stages of categorization, while \textsf{HAMP-II} shows a significantly improved clustering effect. Notably, \textsf{HAMP-II} has successfully distinguished six distinct categories, confirming that \textsf{HAMP-II} achieves faster cluster than \textsf{HAMP-I} due to the second-order mechanism.
 \end{itemize}

\section{Related Work}

\paragraph{Hypergraph Neural Networks.}
Hypergraph learning was first introduced in~\cite{zhou2006learning} as a propagation process on hypergraph. Since then, hypergraph learning~\cite{gao_hypergraph_2022,kim_hypergraph} has developed extensively. 
As an extension of GNNs, \citet{feng2019hypergraph} proposed HGNN that effectively captures high-order interactions by leveraging the vertex-edge-vertex propagation pattern. Further, $\text{HGNN}^+$~\cite{gao2022hgnn} introduced hyperedge groups and adaptive hyperedge group fusion strategy as a general framework for modeling high-order multi-modal/multi-type data correlations. 
Following the spectral theory of hypergraph~\cite{yadati2019hypergcn}, SheafHyperGNN~\cite{NEURIPS2023_27f243af} introduced the sheaf theory to model data relationships in hypergraph more finely. 
% HyperMagNet~\cite{Benko2024HyperMagNetAM} further enhances the representation capability of hypergraph neural networks by introducing the concept of magnetic Laplacian. 
Inspired by Transformers~\cite{vaswani2017attention}, several HNNs~\cite{bai2021hypergraph, chien2022you, Choe_whatsnet} enhanced the feature extraction capability for hypergraph through attention mechanism and centrality for positional encoding. 
Different from using vertex-vertex propagation pattern, several works~\cite{chien2022you,ijcai21UniGNN,wang2022equivariant,yan2024hypergraph} have considered employing multi-phase message passing. 
Among these, UniGNN~\cite{ijcai21UniGNN} presented a unified framework that that facilitates the processing of all hypergraph data through GNNs. In contrast, ED-HNN~\cite{wang2022equivariant} and CoNHD~\cite{zheng2024co} were developed from the perspective of optimization of hyperedge and node potential. Additionally, HDS$^{ode}$~\cite{yan2024hypergraph} adopted control-diffusion ODEs to model the hypergraph dynamic system. By contrast, our method is based on the particle system theory, employing the first-order and second-order systems to understand and design HNNs. 
% For a comprehensive understanding of more works related to hypergraph learning, interested readers are referred to the reviews~\cite{gao_hypergraph_2022,kim_hypergraph}.

\paragraph{Collective Dynamics.}
\citet{watts1998collective} first mathematically defined a small-world network and explained the reasons behind the collective dynamics. Many researchers are interested in how active/passive media made of many interacting agents form complex patterns in mathematical biology and technology. \citet{Battiston_2021} showed that higher-order interactions play a crucial role in understanding these complex patterns. These complex patterns can be seen in animal groups, cell clusters, granular media, and self-organizing particles, as shown in~\cite{carrillo2010particle,holm2006formation,kolokolnikov2013emergent} and other references. In many of these models, the agents move into groups  based on the attractive-repulsive forces~\cite{carrillo2010self,d2006self,carrillo2023radial}. For example, \citet{fang2019emergent} have studied the Cucker-Smale model~\cite{cucker2007emergent} with Rayleigh friction and attractive-repulsive coupling, and \cite{jin2021collective} showed a similar collective phenomenon with stochastic dynamics. Furthermore, the ACMP~\cite{wang2023acmp} was based on Allen-Cahn particle system and incorporated the repulsive force, providing inspiration for our work in hypergraph learning.

\section{Conclusion}
In this paper, we introduce a novel hypergraph message passing framework inspired by particle system theory. We derive both first‐order and second‐order system equations, yielding two distinct models for modeling hypergraph message passing dynamics that capture full hyperedge interactions while mitigating over‐smoothing and heterophily. The proposed models further integrate a stochastic term to model uncertainty in these interactions and can alleviate over‐smoothing in deep layers. By casting HNNs in a physically interpretable paradigm, our model balances high‐order interaction modeling with feature‐diversity preservation, offering both theoretical insights and practical advances for complex system analysis. In future work, we plan to extend the proposed methods to protein‐structure and sequence design for the discovery of novel antibodies and enzymes. 

{
% \small
\bibliographystyle{plainnat}  %plainnat,abbrvnat,unsrtnat,IEEEtran
\bibliography{reference}

\begin{thebibliography}{49}
\providecommand{\natexlab}[1]{#1}
\providecommand{\url}[1]{\texttt{#1}}
\expandafter\ifx\csname urlstyle\endcsname\relax
  \providecommand{\doi}[1]{doi: #1}\else
  \providecommand{\doi}{doi: \begingroup \urlstyle{rm}\Url}\fi

\bibitem[Allen and Cahn(1979)]{allen1979microscopic}
Samuel~M Allen and John~W Cahn.
\newblock A microscopic theory for antiphase boundary motion and its application to antiphase domain coarsening.
\newblock \emph{Acta metallurgica}, 27\penalty0 (6):\penalty0 1085--1095, 1979.

\bibitem[Bai et~al.(2021)Bai, Zhang, and Torr]{bai2021hypergraph}
Song Bai, Feihu Zhang, and Philip~HS Torr.
\newblock Hypergraph convolution and hypergraph attention.
\newblock \emph{Pattern Recognition}, 110:\penalty0 107637, 2021.

\bibitem[Battaglia et~al.(2018)Battaglia, Hamrick, Bapst, Sanchez-Gonzalez, Zambaldi, Malinowski, Tacchetti, Raposo, Santoro, Faulkner, Çaglar Gülçehre, Song, Ballard, Gilmer, Dahl, Vaswani, Allen, Nash, Langston, Dyer, Heess, Wierstra, Kohli, Botvinick, Vinyals, Li, and Pascanu]{battaglia2018relational}
Peter~W. Battaglia, Jessica~B. Hamrick, Victor Bapst, Alvaro Sanchez-Gonzalez, Vinícius~Flores Zambaldi, Mateusz Malinowski, Andrea Tacchetti, David Raposo, Adam Santoro, Ryan Faulkner, Çaglar Gülçehre, H.~Francis Song, Andrew~J. Ballard, Justin Gilmer, George~E. Dahl, Ashish Vaswani, Kelsey~R. Allen, Charles Nash, Victoria Langston, Chris Dyer, Nicolas Heess, Daan Wierstra, Pushmeet Kohli, Matthew Botvinick, Oriol Vinyals, Yujia Li, and Razvan Pascanu.
\newblock Relational inductive biases, deep learning, and graph networks.
\newblock \emph{CoRR}, abs/1806.01261, 2018.

\bibitem[Battiston et~al.(2021)Battiston, Amico, Barrat, Bianconi, Ferraz~de Arruda, Franceschiello, Iacopini, Kéfi, Latora, Moreno, Murray, Peixoto, Vaccarino, and Petri]{Battiston_2021}
Federico Battiston, Enrico Amico, Alain Barrat, Ginestra Bianconi, Guilherme Ferraz~de Arruda, Benedetta Franceschiello, Iacopo Iacopini, Sonia Kéfi, Vito Latora, Yamir Moreno, Micah~M. Murray, Tiago~P. Peixoto, Francesco Vaccarino, and Giovanni Petri.
\newblock The physics of higher-order interactions in complex systems.
\newblock \emph{Nature Physics}, 17\penalty0 (10):\penalty0 1093–1098, October 2021.
\newblock ISSN 1745-2481.

\bibitem[Bazaga et~al.(2024)Bazaga, Lio, and Micklem]{bazaga2024hyperbert}
Adri{\'a}n Bazaga, Pietro Lio, and Gos Micklem.
\newblock {H}yper{BERT}: Mixing hypergraph-aware layers with language models for node classification on text-attributed hypergraphs.
\newblock In \emph{Findings of the Association for Computational Linguistics: EMNLP 2024}, Miami, Florida, USA, November 2024. Association for Computational Linguistics.

\bibitem[Carrillo and Shu(2023)]{carrillo2023radial}
Jos{\'e}~A Carrillo and Ruiwen Shu.
\newblock From radial symmetry to fractal behavior of aggregation equilibria for repulsive--attractive potentials.
\newblock \emph{Calculus of Variations and Partial Differential Equations}, 62\penalty0 (1):\penalty0 28, 2023.

\bibitem[Carrillo et~al.(2010{\natexlab{a}})Carrillo, Fornasier, Toscani, and Vecil]{carrillo2010particle}
Jos{\'e}~A Carrillo, Massimo Fornasier, Giuseppe Toscani, and Francesco Vecil.
\newblock Particle, kinetic, and hydrodynamic models of swarming.
\newblock \emph{Mathematical modeling of collective behavior in socio-economic and life sciences}, pages 297--336, 2010{\natexlab{a}}.

\bibitem[Carrillo et~al.(2010{\natexlab{b}})Carrillo, Klar, Martin, and Tiwari]{carrillo2010self}
Jos{\'e}~A Carrillo, Axel Klar, Stephan Martin, and Sudarshan Tiwari.
\newblock Self-propelled interacting particle systems with roosting force.
\newblock \emph{Mathematical Models and Methods in Applied Sciences}, 20\penalty0 (supp01):\penalty0 1533--1552, 2010{\natexlab{b}}.

\bibitem[Chamberlain et~al.(2021)Chamberlain, Rowbottom, Gorinova, Webb, Rossi, and Bronstein]{chamberlain2021grand}
Benjamin~Paul Chamberlain, James Rowbottom, Maria~I. Gorinova, Stefan~D Webb, Emanuele Rossi, and Michael~M. Bronstein.
\newblock {GRAND}: Graph neural diffusion.
\newblock In \emph{ICML}, 2021.

\bibitem[Chen et~al.(2018)Chen, Rubanova, Bettencourt, and Duvenaud]{chen2018neural}
Ricky~TQ Chen, Yulia Rubanova, Jesse Bettencourt, and David~K Duvenaud.
\newblock Neural ordinary differential equations.
\newblock \emph{NeurIPS}, 31, 2018.

\bibitem[Chien et~al.(2022)Chien, Pan, Peng, and Milenkovic]{chien2022you}
Eli Chien, Chao Pan, Jianhao Peng, and Olgica Milenkovic.
\newblock You are allset: A multiset function framework for hypergraph neural networks.
\newblock In \emph{ICLR}, 2022.

\bibitem[Choe et~al.(2023)Choe, Kim, Yoo, and Shin]{Choe_whatsnet}
Minyoung Choe, Sunwoo Kim, Jaemin Yoo, and Kijung Shin.
\newblock Classification of edge-dependent labels of nodes in hypergraphs.
\newblock In \emph{KDD}, page 298–309, New York, NY, USA, 2023. Association for Computing Machinery.

\bibitem[Cucker and Smale(2007)]{cucker2007emergent}
Felipe Cucker and Steve Smale.
\newblock Emergent behavior in flocks.
\newblock \emph{IEEE Transactions on Automatic Control}, 52\penalty0 (5):\penalty0 852--862, 2007.

\bibitem[Dong et~al.(2020)Dong, Sawin, and Bengio]{dong2020hnhn}
Yihe Dong, Will Sawin, and Yoshua Bengio.
\newblock Hnhn: Hypergraph networks with hyperedge neurons.
\newblock In \emph{ICML}, 2020.

\bibitem[Du and Zhou(2023)]{Linglong_stochastic}
Linglong Du and Xinyun Zhou.
\newblock The stochastic delayed cucker-smale system in a harmonic potential field.
\newblock \emph{Kinetic and Related Models}, 16\penalty0 (1):\penalty0 54--68, 2023.
\newblock ISSN 1937-5093.
\newblock \doi{10.3934/krm.2022022}.

\bibitem[Duta et~al.(2023)Duta, Cassar\`{a}, Silvestri, and Li\'{o}]{NEURIPS2023_27f243af}
Iulia Duta, Giulia Cassar\`{a}, Fabrizio Silvestri, and Pietro Li\'{o}.
\newblock Sheaf hypergraph networks.
\newblock In \emph{NeurIPS}, volume~36, pages 12087--12099. Curran Associates, Inc., 2023.

\bibitem[D’Orsogna et~al.(2006)D’Orsogna, Chuang, Bertozzi, and Chayes]{d2006self}
Maria~R D’Orsogna, Yao-Li Chuang, Andrea~L Bertozzi, and Lincoln~S Chayes.
\newblock Self-propelled particles with soft-core interactions: patterns, stability, and collapse.
\newblock \emph{Physical review letters}, 96\penalty0 (10):\penalty0 104302, 2006.

\bibitem[E(2017)]{weinan2017proposal}
Weinan E.
\newblock A proposal on machine learning via dynamical systems.
\newblock \emph{Communications in Mathematics and Statistics}, 1\penalty0 (5):\penalty0 1--11, 2017.

\bibitem[Fang et~al.(2019)Fang, Ha, and Jin]{fang2019emergent}
Di~Fang, Seung-Yeal Ha, and Shi Jin.
\newblock Emergent behaviors of the {Cucker-Smale} ensemble under attractive-repulsive couplings and rayleigh frictions.
\newblock \emph{Mathematical Models and Methods in Applied Sciences}, 29\penalty0 (07):\penalty0 1349--1385, 2019.

\bibitem[Feng et~al.(2019)Feng, You, Zhang, Ji, and Gao]{feng2019hypergraph}
Yifan Feng, Haoxuan You, Zizhao Zhang, Rongrong Ji, and Yue Gao.
\newblock Hypergraph neural networks.
\newblock In \emph{AAAI}, volume~33, pages 3558--3565, 2019.

\bibitem[Gao et~al.(2022)Gao, Zhang, Lin, Zhao, Du, and Zou]{gao_hypergraph_2022}
Yue Gao, Zizhao Zhang, Haojie Lin, Xibin Zhao, Shaoyi Du, and Changqing Zou.
\newblock Hypergraph {Learning}: {Methods} and {Practices}.
\newblock \emph{TPAMI}, 44\penalty0 (5), May 2022.
\newblock ISSN 1939-3539.

\bibitem[Gao et~al.(2023)Gao, Feng, Ji, and Ji]{gao2022hgnn}
Yue Gao, Yifan Feng, Shuyi Ji, and Rongrong Ji.
\newblock Hgnn+: General hypergraph neural networks.
\newblock \emph{TPAMI}, 45\penalty0 (3):\penalty0 3181--3199, 2023.
\newblock \doi{10.1109/TPAMI.2022.3182052}.

\bibitem[Gilmer et~al.(2017)Gilmer, Schoenholz, Riley, Vinyals, and Dahl]{gilmer2017neural}
Justin Gilmer, Samuel~S Schoenholz, Patrick~F Riley, Oriol Vinyals, and George~E Dahl.
\newblock Neural message passing for quantum chemistry.
\newblock In \emph{ICML}, 2017.

\bibitem[Ha et~al.(2009)Ha, Lee, and Levy]{ha2009emergence}
Seung-Yeal Ha, Kiseop Lee, and Doron Levy.
\newblock Emergence of time-asymptotic flocking in a stochastic cucker-smale system.
\newblock \emph{Communications in Mathematical Sciences}, 7\penalty0 (2), 2009.

\bibitem[Holm and Putkaradze(2006)]{holm2006formation}
Darryl~D Holm and Vakhtang Putkaradze.
\newblock Formation of clumps and patches in self-aggregation of finite-size particles.
\newblock \emph{Physica D: Nonlinear Phenomena}, 220\penalty0 (2):\penalty0 183--196, 2006.

\bibitem[Huang and Yang(2021)]{ijcai21UniGNN}
Jing Huang and Jie Yang.
\newblock Unignn: a unified framework for graph and hypergraph neural networks.
\newblock In \emph{IJCAI}, 2021.

\bibitem[Jin and Shu(2021)]{jin2021collective}
Shi Jin and Ruiwen Shu.
\newblock Collective dynamics of opposing groups with stochastic communication.
\newblock \emph{Vietnam Journal of Mathematics}, 49:\penalty0 619--636, 2021.

\bibitem[Kim et~al.(2024)Kim, Lee, Gao, Antelmi, Polato, and Shin]{kim_hypergraph}
Sunwoo Kim, Soo~Yong Lee, Yue Gao, Alessia Antelmi, Mirko Polato, and Kijung Shin.
\newblock A survey on hypergraph neural networks: An in-depth and step-by-step guide.
\newblock In \emph{KDD}, page 6534–6544, New York, NY, USA, 2024. Association for Computing Machinery.

\bibitem[Kipf and Welling(2017)]{kipf2016semi}
Thomas~N Kipf and Max Welling.
\newblock Semi-supervised classification with graph convolutional networks.
\newblock In \emph{ICLR}, 2017.

\bibitem[Kolokolnikov et~al.(2011)Kolokolnikov, Sun, Uminsky, and Bertozzi]{kolokolnikov2011stability}
Theodore Kolokolnikov, Hui Sun, David Uminsky, and Andrea~L Bertozzi.
\newblock Stability of ring patterns arising from two-dimensional particle interactions.
\newblock \emph{Physical Review E}, 84\penalty0 (1):\penalty0 015203, 2011.

\bibitem[Kolokolnikov et~al.(2013)Kolokolnikov, Carrillo, Bertozzi, Fetecau, and Lewis]{kolokolnikov2013emergent}
Theodore Kolokolnikov, Jos{\'e}~A Carrillo, Andrea Bertozzi, Razvan Fetecau, and Mark Lewis.
\newblock Emergent behaviour in multi-particle systems with non-local interactions, 2013.

\bibitem[Li et~al.(2025)Li, Gu, Wang, Fang, Bai, Zhuang, and Lio]{li2025hypergraph}
Ming Li, Yongchun Gu, Yi~Wang, Yujie Fang, Lu~Bai, Xiaosheng Zhuang, and Pietro Lio.
\newblock When hypergraph meets heterophily: New benchmark datasets and baseline.
\newblock In \emph{AAAI}, pages 18377--18384, 2025.

\bibitem[Motsch and Tadmor(2014)]{motsch2014heterophilious}
Sebastien Motsch and Eitan Tadmor.
\newblock Heterophilious dynamics enhances consensus.
\newblock \emph{SIAM Review}, 56\penalty0 (4):\penalty0 577--621, 2014.

\bibitem[Nt and Maehara(2019)]{nt2019revisiting}
Hoang Nt and Takanori Maehara.
\newblock Revisiting graph neural networks: All we have is low-pass filters.
\newblock \emph{arXiv preprint arXiv:1905.09550}, 2019.

\bibitem[Oono and Suzuki(2019)]{oono2019graph}
Kenta Oono and Taiji Suzuki.
\newblock Graph neural networks exponentially lose expressive power for node classification.
\newblock In \emph{ICLR}, 2019.

\bibitem[Prokopchik et~al.(2022)Prokopchik, Benson, and Tudisco]{prokopchik2022nonlinear}
Konstantin Prokopchik, Austin~R Benson, and Francesco Tudisco.
\newblock Nonlinear feature diffusion on hypergraphs.
\newblock In \emph{ICML}, pages 17945--17958. PMLR, 2022.

\bibitem[Rayleigh(1894)]{rayleigh1894jws}
Loan Rayleigh.
\newblock \emph{JWS, The Theory of Sound, vol. 1}.
\newblock Macmillan, London (reprinted Dover, New York, 1945), 1894.

\bibitem[van~der Maaten and Hinton(2008)]{vandermaaten08a}
Laurens van~der Maaten and Geoffrey Hinton.
\newblock Visualizing data using t-sne.
\newblock \emph{Journal of Machine Learning Research}, 9\penalty0 (86):\penalty0 2579--2605, 2008.

\bibitem[Vaswani et~al.(2017)Vaswani, Shazeer, Parmar, Uszkoreit, Jones, Gomez, Kaiser, and Polosukhin]{vaswani2017attention}
Ashish Vaswani, Noam Shazeer, Niki Parmar, Jakob Uszkoreit, Llion Jones, Aidan~N. Gomez, \L{}ukasz Kaiser, and Illia Polosukhin.
\newblock Attention is all you need.
\newblock In \emph{NeurIPS}, pages 6000--6010, 2017.

\bibitem[Vi{\~n}as et~al.(2023)Vi{\~n}as, Joshi, Georgiev, Lin, Dumitrascu, Gamazon, and Li{\`o}]{vinas2023hypergraph}
Ramon Vi{\~n}as, Chaitanya~K Joshi, Dobrik Georgiev, Phillip Lin, Bianca Dumitrascu, Eric~R Gamazon, and Pietro Li{\`o}.
\newblock Hypergraph factorization for multi-tissue gene expression imputation.
\newblock \emph{Nature Machine Intelligence}, pages 1--15, 2023.

\bibitem[Wang et~al.(2024)Wang, Liu, Liu, Wang, Medya, and Yu]{wang2024ugnn}
Fangxin Wang, Yuqing Liu, Kay Liu, Yibo Wang, Sourav Medya, and Philip~S. Yu.
\newblock Uncertainty in graph neural networks: A survey.
\newblock \emph{Transactions on Machine Learning Research}, 2024.

\bibitem[Wang et~al.(2023{\natexlab{a}})Wang, Yang, Liu, Wang, and Li]{wang2022equivariant}
Peihao Wang, Shenghao Yang, Yunyu Liu, Zhangyang Wang, and Pan Li.
\newblock Equivariant hypergraph diffusion neural operators.
\newblock In \emph{ICLR}, 2023{\natexlab{a}}.

\bibitem[Wang et~al.(2023{\natexlab{b}})Wang, Yi, Liu, Wang, and Jin]{wang2023acmp}
Yuelin Wang, Kai Yi, Xinliang Liu, Yu~Guang Wang, and Shi Jin.
\newblock {ACMP}: Allen-cahn message passing with attractive and repulsive forces for graph neural networks.
\newblock In \emph{ICLR}, 2023{\natexlab{b}}.

\bibitem[Watts and Strogatz(1998)]{watts1998collective}
Duncan~J Watts and Steven~H Strogatz.
\newblock Collective dynamics of ‘small-world’networks.
\newblock \emph{Nature}, 393\penalty0 (6684):\penalty0 440--442, 1998.

\bibitem[Wu et~al.(2025)Wu, Wu, Liu, Liu, Shao, and Wang]{wu2024se3set}
Hongfei Wu, Lijun Wu, Guoqing Liu, Zhirong Liu, Bin Shao, and Zun Wang.
\newblock Se3set: Harnessing equivariant hypergraph neural networks for molecular representation learning.
\newblock \emph{Transactions on Machine Learning Research}, 2025.

\bibitem[Yadati et~al.(2019)Yadati, Nimishakavi, Yadav, Nitin, Louis, and Talukdar]{yadati2019hypergcn}
Naganand Yadati, Madhav Nimishakavi, Prateek Yadav, Vikram Nitin, Anand Louis, and Partha Talukdar.
\newblock Hypergcn: A new method for training graph convolutional networks on hypergraphs.
\newblock In \emph{NeurIPS}, 2019.

\bibitem[Yan et~al.(2024)Yan, Feng, Ying, and Gao]{yan2024hypergraph}
Jielong Yan, Yifan Feng, Shihui Ying, and Yue Gao.
\newblock Hypergraph dynamic system.
\newblock In \emph{ICLR}, 2024.

\bibitem[Zheng and Worring(2024)]{zheng2024co}
Yijia Zheng and Marcel Worring.
\newblock Co-representation neural hypergraph diffusion for edge-dependent node classification.
\newblock \emph{arXiv preprint arXiv:2405.14286}, 2024.

\bibitem[Zhou et~al.(2006)Zhou, Huang, and Sch{\"o}lkopf]{zhou2006learning}
Dengyong Zhou, Jiayuan Huang, and Bernhard Sch{\"o}lkopf.
\newblock Learning with hypergraphs: Clustering, classification, and embedding.
\newblock In \emph{NeurIPS}, volume~19, 2006.

\end{thebibliography}
}

%%%%%%%%%%%%%%%%%%%%%%%%%%%%%%%%%%%%%%%%%%%%%%%%%%%%%%%%%%%%

%%%%%%%%%%%%%%%%%%%%%%%%%%%%%%%%%%%%%%%%%%%%%%%%%%%%%%%%%%%%
\newpage
\appendix

% \section*{Appendix}
% We include several appendices with algorithms, theoretical proof, and experiment details. First, we detail on two algorithms for \textsf{HAMP} framework in Appendix~\ref{alg}, with complexity analysis and limitations discussion. Next, we provide the theoretical results and proofs in Appendix~\ref{proof}. Finally, we detail dataset details, additional ablation studies, time-memory tradeoff analysis, and hyperparameters in Appendix~\ref{ex_details}.

\section{Algorithms}\label{alg}

\paragraph{Complexity Analysis.}
Here, we analyze the computational complexity of one layer in \textsf{HAMP-I} and \textsf{HAMP-II}. Analytically, the time complexity is 
$\mathcal{O}\left(|\mathcal{V}| |\mathcal{E}| c^2 + |\mathcal{V}|c\right)$, where $|\mathcal{V}|$, $|\mathcal{E}|$ and $c$ are the number of nodes, number of hyperedges and number of hidden dimension, respectively.
However, the incidence matrix $\mathbf{H}$ is a sparse matrix, so the time complexity is $\mathcal{O}\left((tr(\mathbf{D}_v) + tr(\mathbf{D}_e)) c^2 + |\mathcal{V}|c \right)$, 
where $tr(\mathbf{D}_v)$ is the sum of the degrees of all nodes and $tr(\mathbf{D}_e)$ is the sum of the number of nodes contained in all hyperedges.
The detailed process of \textsf{HAMP-I} and \textsf{HAMP-II} are shown in Algorithm \ref{algorithm1} and Algorithm \ref{algorithm2}.

\begin{algorithm}[h]
\caption{The \textsf{HAMP-I} Algorithm for Hypergraph Node Classification.}\label{algorithm1}
\begin{algorithmic}[1]
\STATE \textbf{Input}: the incidence matrix $\mathbf{H}$, the node feature $\mathbf{X}$, and the node labels $\mathbf{Y}$\,.
\STATE \textbf{Output}: the model prediction accuracy\,.
\STATE \textbf{Initialization}: the time $T$, the step size $\tau$ and all parameters of model\,.
\WHILE{not converged}
    \STATE Node feature mapping $\mathbf{X} = \text{Linear}_{\text{map}}(\mathbf{X})$\,;   
    \STATE Set the initial time $t=0$, the initial node representation $\mathbf{X}(0) = \mathbf{X}$\,;
    \WHILE{$t \le T$}
        \STATE Message passing from $\mathcal{V}$ to $\mathcal{E}$: 
        $\mathbf{X}_{\mathcal{V} \to \mathcal{E}}(t) = \Phi_{1} (\mathbf{X}(t), \mathbf{H})$\,;
        \STATE Message passing from $\mathcal{E}$ to $\mathcal{V}$: 
        $\mathbf{X}_{\mathcal{E} \to \mathcal{V}}(t) = \Psi\left(\mathbf{X}(t), \Phi_{2} \left( \mathbf{X}_{\mathcal{V} \to \mathcal{E}}(t), \mathbf{H} \right) \right)$\,;
        \STATE Compute particle dynamics  as \\
        $\mathbf{X}(t+\tau) = \mathbf{X}(t) + \tau \sigma \left ( \underbrace{\mathbf{X}_{\mathcal{E} \to \mathcal{V}}(t) - \omega \mathbf{X}(t)}_{\text {Interaction Force}} + \underbrace{\delta f_d(\mathbf{X}(t))}_{\text {Allen-Cahn Force}} + \underbrace{\epsilon \mathbf{B}(t)}_{\text {Noise}} + \beta \mathbf{X}(0) \right )$\,;
        \STATE Updata $t = t +\tau$\,;
    \ENDWHILE
    \STATE Input the node representation into the classifier $\mathbf{X}^{out} = \text{MLP} (\mathbf{X}(T))$;
    \STATE Compute the model prediction labels $\mathbf{\widehat{Y}} = \text{Softmax}(\mathbf{X}^{out})$ and compute the loss function\,;
    \STATE Update all parameter by back propagation using the  Adam optimizer\,;
\ENDWHILE
\end{algorithmic}
\end{algorithm}

\begin{algorithm}[h]
\caption{The \textsf{HAMP-II} Algorithm for Hypergraph Node Classification.}\label{algorithm2}
\begin{algorithmic}[1]
\STATE \textbf{Input}: the incidence matrix $\mathbf{H}$, the node feature $\mathbf{X}$, and the node labels $\mathbf{Y}$\,.
\STATE \textbf{Output}: the model prediction accuracy\,.
\STATE \textbf{Initialization}: the time $T$, the step size $\tau$ and all parameters of model\,.
\WHILE{not converged}
    \STATE Node feature mapping $\mathbf{X} = \text{Linear}_{\text{map}}(\mathbf{X})$\,; 
    \STATE Set the initial time $t=0$, the initial node representation $\mathbf{X}(0) = \mathbf{X}$\,;
    \STATE Set the initial velocity $\mathbf{V}(0) = \text{Linear}(\mathbf{X}(0)) - \mathbf{X}(0)$\,;
    \WHILE{$t \le T$}
        \STATE Message passing from $\mathcal{V}$ to $\mathcal{E}$: 
        $\mathbf{V}_{\mathcal{V} \to \mathcal{E}}(t) = \Phi_{1} (\mathbf{V}(t), \mathbf{H})$\,;
        \STATE Message passing from $\mathcal{E}$ to $\mathcal{V}$: 
        $\mathbf{V}_{\mathcal{E} \to \mathcal{V}}(t) = \Psi\left(\mathbf{V}(t), \Phi_{2} \left( \mathbf{V}_{\mathcal{V} \to \mathcal{E}}(t), \mathbf{H} \right) \right)$\,;
        \STATE Compute the velocity of particle dynamics system  as \\
        $\mathbf{V}(t+\tau) = \mathbf{V}(t) + $
        $\tau \sigma \left( \underbrace{\mathbf{V}_{\mathcal{E} \to \mathcal{V}}(t) - \omega \mathbf{V}(t)}_{\text {Interaction Force}} + \underbrace{\delta f_d(\mathbf{V}(t))}_{\text {Allen-Cahn Force}} + \underbrace{\epsilon \mathbf{B}(t)}_{\text {Noise}} + \beta \mathbf{V}(0) \right)$\,;
        \STATE Compute the representation by 
        $\mathbf{X}(t+\tau) = \mathbf{X}(t) + \tau\mathbf{V}(t+\tau)$\,;
        \STATE Updata $t = t +\tau$\,;
    \ENDWHILE
    \STATE Input the node representation into the classifier $\mathbf{X}^{out} = \text{MLP} (\mathbf{X}(T))$;
    \STATE Compute the model prediction labels $\mathbf{\widehat{Y}} = \text{Softmax}(\mathbf{X}^{out})$ and compute the loss function\,;
    \STATE Update all parameter by back propagation using the  Adam optimizer\,;
\ENDWHILE
\end{algorithmic}
\end{algorithm}

\paragraph{Limitations Discussion.}\label{lim_fut}
Our particle dynamics-based hypergraph message passing framework assumes a static hypergraph topology. While this assumption is valid for social and biological hypergraphs with slowly evolving interactions, it may not hold in highly dynamic scenarios like financial transaction networks, where hyperedge topologies change abruptly. Consequently, effectively modeling temporal hypergraphs with evolving structures remains an open challenge.

\section{Theoretical Results and Proof}\label{proof}

Technically, we suppose there exists $\{f_\beta^e\}$ such that $\mathcal{I}=\{1,\cdots,N\}$ can be divided into two disjoint groups with $N_1, N_2$ particles respectively: $f_{\beta}(h_{i,j}^e)\ge 0,$ for $\{i,j\}\in \mathcal{I}_1$ or $\mathcal{I}_2$ and $ f_{\beta}(h_{i,j}^e)\le 0,$ otherwise. 
We designate 
\begin{equation}
\{x^{(1)}_i\}:=\{\mathbf{x}_i | i\in\mathcal{I}_1\}, \quad \{x^{(2)}_j\}:=\{\mathbf{x}_j | j\in\mathcal{I}_2\}. 
\end{equation}
The model is channel-wise, hence we use $x$ instead of $\mathbf{x}$ in the proof.

First, let's define the relevant notation,

The mean value: 
\begin{equation}
\Bar{x} := \frac{1}{N}\sum\limits_{i=1}^N x_i.
\end{equation}

The deviation values: 
\begin{equation}
\hat{x}_i := x_i - \Bar{x}.
\end{equation}

The variance of values within each group:
\begin{equation}
\text{var}({x}^{(1)}) = \frac{1}{N_1} \sum (\hat{x}^{(1)}_i)^2, 
\quad 
\text{var}({x}^{(2)}) = \frac{1}{N_2} \sum (\hat{x}^{(2)}_i)^2.
\end{equation}

The second moments: 
\begin{equation}
M_2({x}^{(1)}):= \sum \limits_{i=1}^{N_1} ({x}^{(1)}_i)^2, 
\quad
M_2({x}^{(2)}):= \sum \limits_{i=1}^{N_2} ({x}^{(2)}_i)^2.
\end{equation}

And others: 
\begin{equation}
\widehat{M}_2:= M_2(\hat{x}^{(1)})+M_2(\hat{x}^{(2)}) =\text{var}(x^{(1)})+\text{var}(x^{(2)}). 
\end{equation}

\begin{equation}
\psi_i^{e,\pm}:= \sum\limits_{j\in e} h_{i,j}^{e,\pm}, \quad
\psi_i^{\pm}:= \sum\limits_{e\in\mathcal{E}}\sum\limits_{j\in e} h_{i,j}^{e,\pm}.
\end{equation}

\begin{equation}
k:= \max\limits_i\{|\mathcal{E}(i)|\}.
\end{equation}

\begin{equation}
D_e^- := \max\limits_k\{\psi_k^{e,-}\},\,D^- := \max\limits_e\{D_e^-\}.
\end{equation}

\begin{equation}
D_2^- := \max\limits_e\{\|\psi^{e,-}\|_2\}.
\end{equation}

Technically, we set $N_1 = N_2 := N_0$. This assumption is that $N_1$ is comparable to $N_2,$ i.e., there exists a positive constant $\kappa$ satisfying $\frac{1}{\kappa}N_1\le N_2 \le \kappa N_1$.

% We set $\{x^{(1)}_i\}:=\{\mathbf{x}_i | i\in\mathcal{I}_1\}$ and $\{x^{(2)}_j\}:=\{\mathbf{x}_j | j\in\mathcal{I}_2\}.$ 
We can rewrite \eqref{eq:1st order} as
    \begin{equation}\label{eq: bi}
    \left\{
    \begin{aligned}
        \Dot{u}^{(1)}_i &= \frac{1}{N_1}\sum\limits_{e\in\mathcal{E}(i)}\sum\limits_{i^\prime=1}^{N_1}h_{i,i^\prime}^{e,+}(x^{(1)}_{i^\prime}-x^{(1)}_i)
            -  \frac{1}{N_2}\sum\limits_{e\in\mathcal{E}(i)}\sum\limits_{j=1}^{N_2}h_{i,j}^{e,-}(x^{(2)}_{j}-x^{(1)}_i)
            + \delta x^{(1)}_i(1-(x^{(1)}_i)^2)\\
        \Dot{v}^{(2)}_j &= \frac{1}{N_2}\sum\limits_{e\in\mathcal{E}(j)}\sum\limits_{j^\prime=1}^{N_2}h_{j,j^\prime}^{e,+}(x^{(2)}_{j^\prime}-x^{(2)}_j)
            -  \frac{1}{N_1}\sum\limits_{e\in\mathcal{E}(j)}\sum\limits_{i=1}^{N_1}h_{i,j}^{e,-}(x^{(1)}_{i}-x^{(2)}_j)
            + \delta x^{(2)}_j(1-(x^{(2)}_j)^2).
    \end{aligned}
    \right.
\end{equation}
Set the matrix $A^e$ as
\begin{equation}\label{eq:matrix3.1}
    A_{i,j}^e=\left\{
    \begin{aligned}
        h_{i,i}^e\\
        - h_{i,j}^e,
    \end{aligned}\right.
\end{equation}
and and designate $C^A := \min\limits_{e\in\mathcal{E}}\{F(A^e)\},$ where $F(A^e)$ is the \textit{Fiedler number} of $A^e$.

% \subsection{Results on first-order system}

\begin{lemma}[$L_2$ estimate for $M_2$]\label{lem: M2 estimate}
There exists a positive constant $M_2^\infty$ such that 
\begin{equation}
  \sup\limits_{0\le t<\infty}M_2(t) \le M_2^\infty \le \infty. 
\end{equation}
\end{lemma}
\textit{Proof.} Note that $h^{e,\pm}_{i,j}=h^{e,\pm}_{j,i}$, then \\
\begin{equation}
\begin{aligned}
    \frac{\mathrm{d}}{\mathrm{d}t}M_2(x^{(1)}) &= \frac{2}{N_1}\sum_{i=1}^{N_1}x^{(1)}_i \Dot{x}^{(1)}_i    \\
    &= \frac{2}{N_1}\sum\limits_{i=1}^{N_1}\sum\limits_{e\in\mathcal{E}(i)}\sum\limits_{i^\prime \in e} h_{i,i^\prime}^{e,+}(x^{(1)}_{i^\prime}-x^{(1)}_i) x^{(1)}_i
    -\frac{2}{N_1}\sum\limits_{i=1}^{N_1}\sum\limits_{e\in\mathcal{E}(i)}\sum\limits_{j \in e} h_{i,j}^{e,-}(x^{(2)}_j-x^{(1)}_i) x^{(1)}_i\\
    &\qquad+\frac{2\delta}{N_1}\sum\limits_{i=1}^{N_1}(x^{(1)}_i)^2(1-(x^{(1)}_i)^2)\\
    &= \frac{2}{N_1}\sum\limits_{e\in\mathcal{E}}\sum\limits_{i,i^\prime \in e} h_{i,i^\prime}^{e,+}(x^{(1)}_{i^\prime}-x^{(1)}_i)x^{(1)}_i
    -\frac{2}{N_1}\sum\limits_{e\in\mathcal{E}}\sum\limits_{i,j \in e} h_{i,j}^{e,-}(x^{(2)}_j-x^{(1)}_i) x^{(1)}_i \\
    &\qquad +\frac{2\delta}{N_1}\sum\limits_{i=1}^{N_1}(x^{(1)}_i)^2(1-(x^{(1)}_i)^2) 
    \\
    &=-\frac{1}{N_1}\sum\limits_{e\in \mathcal{E}}\sum\limits_{i,i^\prime\in e} h^{e,+}_{i,i^\prime}(x^{(1)}_{i^\prime} -x^{(1)}_{i})^2 - \frac{2}{N_1}\sum\limits_{e\in\mathcal{E}}\sum\limits_{i,j \in e} h_{i,j}^{e,-}(x^{(2)}_j-x^{(1)}_i) x^{(1)}_i  \\
    &\qquad +\frac{2\delta}{N_1}\sum\limits_{i=1}^{N_1}(x^{(1)}_i)^2(1-(x^{(1)}_i)^2).
    \end{aligned}
\end{equation}
Similarly,
\begin{equation}
\begin{aligned}
    \frac{\mathrm{d}}{\mathrm{d}t}M_2(x^{(2)}) &= \frac{2}{N_2}\sum_{j=1}^{N_2}x^{(2)}_j \Dot{x}^{(2)}_j\\
    % &= \frac{2}{N_2}\sum\limits_{e\in\mathcal{E}}\sum\limits_{j,j^\prime\in e} h_{j,j^\prime}^{e,+}(x^{(2)}_{j^\prime}-x^{(2)}_j)x^{(2)}_j -\frac{2} {N_2}\sum\limits_{e\in\mathcal{E}}\sum\limits_{j,i \in e} h_{j,i}^{e,-}(x^{(1)}_i-x^{(2)}_j)x^{(2)}_j \\
    % &\qquad +\frac{2\delta}{N_2}\sum\limits_{j=1}^{N_2}(x^{(2)}_j)^2(1-(x^{(2)}_j)^2).
    % \\
    &= - \frac{1}{N_2}\sum\limits_{e\in \mathcal{E}}\sum\limits_{j,j^\prime\in e} h^{e,+}_{j,j^\prime}(x^{(2)}_{j^\prime}-x^{(2)}_j)^2 -\frac{2} {N_2}\sum\limits_{e\in\mathcal{E}}\sum\limits_{j,i \in e} h_{j,i}^{e,-}(x^{(1)}_i-x^{(2)}_j)x^{(2)}_j  \\
    &\qquad +\frac{2\delta}{N_2}\sum\limits_{j=1}^{N_2}(x^{(2)}_j)^2(1-(x^{(2)}_j)^2).
    \end{aligned}
\end{equation}

% Sum the $M_2(x^{(1)})$ and $M_2(x^{(2)})$.
Define the total second moment $M_2:= M_2(x^{(1)})+M_2(x^{(2)})$. Its time derivative is:
\begin{equation}
\frac{\mathrm{d}}{\mathrm{d}t}M_2 = \frac{\mathrm{d}}{\mathrm{d}t}M_2(x^{(1)})+ \frac{\mathrm{d}}{\mathrm{d}t}M_2(x^{(2)}).
\end{equation}
% Substituting the expressions, focus on the cross terms involving interactions between groups. Note that $N_1 = N_2 := N_0$ and $h^{e,\pm}_{i,j}=h^{e,\pm}_{j,i}$, relabel indices $(j,i)\rightarrow (i,j)$ in the second term: 
% \begin{equation}
% \begin{aligned}
% &- \frac{2}{N_1}\sum\limits_{e\in\mathcal{E}}\sum\limits_{i,j \in e} h_{i,j}^{e,-}(x^{(2)}_j-x^{(1)}_i)x^{(1)}_i -\frac{2}{N_2}\sum\limits_{e\in\mathcal{E}}\sum\limits_{j,i \in e} h_{j,i}^{e,-}(x^{(1)}_i-x^{(2)}_j)x^{(2)}_j \\
% % &= - \frac{2}{N_1}\sum\limits_{e\in\mathcal{E}}\sum\limits_{i,j \in e} h_{i,j}^{e,-}(x^{(2)}_j-x^{(1)}_i)x^{(1)}_i -\frac{2}{N_2}\sum\limits_{e\in\mathcal{E}}\sum\limits_{i,j \in e} h_{i,j}^{e,-}(x^{(1)}_j-x^{(2)}_i)x^{(2)}_j \\
% % &=- \frac{2}{N_1}\sum\limits_{e\in\mathcal{E}}\sum\limits_{i,j \in e} h_{i,j}^{e,-}\left(x^{(1)}_i(x^{(2)}_j - x^{(1)}_i) + \frac{N_1}{N_2} x^{(2)}_j(x^{(2)}_j- x^{(1)}_i) \right) \\
% =& \frac{1}{N_1}\sum\limits_{e\in\mathcal{E}}\sum\limits_{i,j \in e} h_{i,j}^{e,-} (x^{(2)}_j - x^{(1)}_i)^2 + \frac{1}{N_2}\sum\limits_{e\in\mathcal{E}}\sum\limits_{j,i \in e} h_{j,i}^{e,-} (x^{(1)}_i - x^{(2)}_j)^2.
% \end{aligned}
% \end{equation}

By discarding the non-positive squared terms (first two sums), we obtain the inequality:
\begin{equation}
    \begin{aligned}
        \frac{\mathrm{d}}{\mathrm{d}t}M_2 
        \le &- \frac{2}{N_1}\sum\limits_{e\in\mathcal{E}}\sum\limits_{i,j \in e} h_{i,j}^{e,-}(x^{(2)}_j-x^{(1)}_i)x^{(1)}_i -\frac{2}{N_2}\sum\limits_{e\in\mathcal{E}}\sum\limits_{j,i \in e} h_{j,i}^{e,-}(x^{(1)}_i-x^{(2)}_j)x^{(2)}_j 
        \\
        & \quad + \frac{2\delta}{N_1}\sum\limits_{i=1}^{N_1}(x^{(1)}_i)^2 (1- (x^{(1)}_i)^2) + \frac{2\delta}{N_2}\sum\limits_{i=1}^{N_2}(x^{(2)}_i)^2 (1- (x^{(2)}_i)^2).
    \end{aligned}
\end{equation}

By the Cauchy-Schwarz inequality, 
\begin{equation*}
\begin{aligned}
    &(M_2^{(1)})^2 = \left(\sum_{i=1}^{N_1} (x^{(1)}_i)^2\right)^2\le N_1\sum\limits_{i=1}^{N_1}(x^{(1)}_i)^4, \\
    &(M_2^{(2)})^2 =\left(\sum_{i=1}^{N_2} (x^{(2)}_i)^2\right)^2\le N_2\sum\limits_{i=1}^{N_2}(x^{(2)}_i)^4, \\
    &(x^{(1)}_i-x^{(2)}_j)^2\le 2 ((x^{(1)}_i)^2+(x^{(2)}_j)^2).
\end{aligned}
\end{equation*}

Then we have
\begin{equation}
    \begin{aligned}
    &\frac{2\delta}{N_1}\sum\limits_{i=1}^{N_1}(x^{(1)}_i)^2 \left(1- (x^{(1)}_i)^2\right) + \frac{2\delta}{N_2}\sum\limits_{i=1}^{N_2} (x^{(2)}_i)^2 \left(1- (x^{(2)}_i)^2\right) \\
    =& \frac{2\delta}{N_1}\sum\limits_{i=1}^{N_1} (x^{(1)}_i)^2 - \frac{2\delta}{N_1}\sum\limits_{i=1}^{N_1}(x^{(1)}_i)^4 + \frac{2\delta}{N_2}\sum\limits_{i=1}^{N_2}(x^{(2)}_i)^2 - \frac{2\delta}{N_2}\sum\limits_{i=1}^{N_2}(x^{(2)}_i)^4 \\
    \le & \frac{2\delta}{N_1}\sum\limits_{i=1}^{N_1} (x^{(1)}_i)^2 - \frac{2\delta}{(N_1)^2} \left(\sum\limits_{i=1}^{N_1}(x^{(1)}_i)^2 \right)^2 + \frac{2\delta}{N_2}\sum\limits_{i=1}^{N_2} (x^{(2)}_i)^2 - \frac{2\delta}{(N_2)^2} \left(\sum\limits_{i=1}^{N_2}(x^{(2)}_i)^2\right)^2  
    \\
     = & 2\delta \left(\frac{M_2^{(1)}}{N_1} + \frac{M_2^{(2)}}{N_2}\right) - 2\delta \left(\left(\frac{M_2^{(1)}}{N_1}\right)^2 + \left(\frac{M_2^{(2)}}{N_2}\right)^2\right) 
    \\
    \le & 2\delta \left(\frac{M_2^{(1)}}{N_1} + \frac{M_2^{(2)}}{N_2}\right) - \delta \left(\frac{M_2^{(1)}}{N_1} + \frac{M_2^{(2)}}{N_2}\right)^2 
    \\
    % = & 2\delta \left(\frac{M_2}{N}\right) - \delta \left(\frac{M_2}{N}\right)^2 
    % \\
    \le & \frac{2\delta}{N' } M_2 - \frac{\delta}{N''} (M_2)^2,
    \end{aligned}
\end{equation}
where $N'=\min\{N_1,N_2\}$ and $N''=\max\{N_1,N_2\}$.

So, we have
\begin{equation}
    \begin{aligned}
        \frac{\mathrm{d}}{\mathrm{d}t}M_2
        \le &\frac{D^-}{N_1}\sum\limits_{e\in\mathcal{E}}\sum\limits_{i,j \in e} \left((x^{(2)}_j-x^{(1)}_i)^2+(x^{(1)}_i)^2\right) +\frac{D^-}{N_2}\sum\limits_{e\in\mathcal{E}}\sum\limits_{j,i \in e} \left((x^{(1)}_i-x^{(2)}_j)^2+(x^{(2)}_j)^2\right) \\
        &\qquad + \frac{2\delta}{N_1}\sum\limits_{i=1}^{N_1}(x^{(1)}_i)^2 \left(1- (x^{(1)}_i)^2\right) + \frac{2\delta}{N_2}\sum\limits_{i=1}^{N_2}(x^{(2)}_i)^2 \left(1- (x^{(2)}_i)^2\right)\\
        \le &\frac{D^-}{N_1}\sum\limits_{e\in\mathcal{E}}\sum\limits_{i,j \in e} \left(3(x^{(1)}_i)^2+2(x^{(2)}_j)^2\right) 
        +\frac{D^-}{N_2}\sum\limits_{e\in\mathcal{E}}\sum\limits_{i,j \in e} \left(2(x^{(1)}_i)^2+3(x^{(2)}_j)^2\right) 
        + \frac{2\delta}{N' } M_2 -\frac{\delta}{N'' } M_2^2.
    \end{aligned}
\end{equation}

These relations yield a Riccati-type differential inequality:
\begin{equation}
\begin{aligned}
    \frac{\mathrm{d}}{\mathrm{d}t}M_2 \le&
    5D^-\max\bigl\{\frac{1}{N_1},\frac{1}{N_2}\bigr\}\sum\limits_{e\in\mathcal{E}}\sum\limits_{i,j \in e} \left((x^{(1)}_i)^2+(x^{(2)}_j)^2\right) + 2\delta M_2 -\delta (M_2)^2\\
    \le& \frac{5D^- k}{N'} M_2 + \frac{2\delta}{N'} M_2 -\frac{\delta}{N''} (M_2)^2\\
    \le &\frac{5D^- k+2\delta}{N'}M_2 -\frac{\delta}{N''}(M_2)^2.
\end{aligned}
\end{equation}

Let $y$ be a solution of the following ODE:
\begin{equation}\label{eq:y'}
    y^\prime = a y - b y^2.
\end{equation}
Then, by phase line analysis, the solution $y(t)$ to \eqref{eq:y'} satisfies 
\begin{equation}\label{2.7Fang}
    M_2(t)\le y(t)\le\max \left\{\frac{a}{b}, M_2(0)\right\} = \max \left\{\frac{(5D^- k+2\delta)N''}{\delta N'}, M_2(0)\right\} =: M_2^\infty.
\end{equation}
which yields the desired estimate.
\bs

\begin{proposition}\label{prop: bounded}
    For \eqref{eq:1st order}, the distance of the centers of the two clusters is finite. 
\end{proposition}

\textit{Proof of Propostion \ref{prop: bounded}.}
By Lemma \ref{lem: M2 estimate}
\begin{equation}
\begin{aligned}
    \| \Bar{x}^{(1)}-\Bar{x}^{(2)} \| &= \Bigl\|\frac{1}{N_1}\sum\limits_{i=1}^{N_1}x^{(1)}_i - \frac{1}{N_2}\sum\limits_{j=1}^{N_2}x^{(2)}_j\Bigr\|  \\[1mm]
    &\le \sqrt{2}\sqrt{\frac{1}{N_1}\sum\limits_{i=1}^{N_1} (x^{(1)}_i)^2 + \frac{1}{N_2}\sum\limits_{j=1}^{N_2}(x^{(2)}_j)^2}\\[1mm]
    &\le \sqrt{2}\sqrt{M_2^\infty},
\end{aligned}
\end{equation}
where the first inequality used the Cauchy-Schwarz inequality.
\bs

\begin{lemma}\label{lem: L2 estimate for central dist}
    Let $u,v$ be the solution to \eqref{eq: bi}. Then $\|\Bar{x}^{(1)}-\Bar{x}^{(2)}\|^2$ satisfies
    \begin{equation}
        \frac{1}{2}\frac{\mathrm{d}}{\mathrm{d}t}\|\Bar{x}^{(1)}-\Bar{x}^{(2)}\|^2 \ge \left(\frac{2c_m}{N_0} - c_1\right) \|\Bar{x}^{(1)}-\Bar{x}^{(2)}\|^2 
            - \frac{4(D_2^-)^2}{c_1 N_0}k \widehat{M}_2.
    \end{equation}
\end{lemma}
\textit{Proof.} 
The time evolution of $\Bar{x}^{(1)}$ is given by
\begin{equation}\label{eq:4.7 diff of mean}
\begin{aligned}
    \Dot{\Bar{x}}^{(1)} &= \frac{1}{N_1}\sum\limits_{i=1}^{N_1} \sum\limits_{i^\prime\in\mathcal{E}(i)} h_{i,i^\prime}^{e,+}(x^{(1)}_{i^\prime}-x^{(1)}_i)
        - \frac{1}{N_1}\sum\limits_{i=1}^{N_1} \sum\limits_{j\in\mathcal{E}(i)} h_{j,i}^{e,-}(x^{(2)}_j-x^{(1)}_i)\\
        &= -\frac{1}{N_1} \sum\limits_{e\in\mathcal{E}}\sum\limits_{i,j\in e} h_{i,j}^{e,-}(x^{(2)}_j-x^{(1)}_i)\\
        &= -\frac{1}{N_1} \sum\limits_{e\in\mathcal{E}}\psi_j^{e,-}x^{(2)}_j + \frac{1}{N_1}\sum\limits_{e\in\mathcal{E}}\psi_i^{e,-}x^{(1)}_i\\
        &= \sum\limits_{e\in\mathcal{E}}\left(-\frac{1}{N_1}\psi_j^{e,-}x^{(2)}_j + \frac{1}{N_1}\psi_i^{e,-}x^{(1)}_i\right),
\end{aligned}
\end{equation}
where the first equality uses the relation $\sum\limits_{i=1}^{N_1}\hat{x}^{(1)}_i = 0$. 

Then we have
% \begin{equation}
\begin{align}
    &\frac{1}{2}\frac{\mathrm{d}}{\mathrm{d}t}\|\Bar{x}^{(1)}-\Bar{x}^{(2)}\|^2 \\ 
    =& (\Bar{x}^{(1)}-\Bar{x}^{(2)})(\Dot{\Bar{x}}^{(1)}-\Dot{\Bar{x}}^{(2)})\\
    =& (\Bar{x}^{(1)}-\Bar{x}^{(2)}) \left[\sum\limits_{e\in\mathcal{E}}\left(-\frac{1}{N_1}\psi_j^{e,-}x^{(2)}_j + \frac{1}{N_1}\psi_i^{e,-}x^{(1)}_i\right)
    - \sum\limits_{e\in\mathcal{E}}\left(-\frac{1}{N_2}\psi_i^{e,-}x^{(1)}_i + \frac{1}{N_2}\psi_j^{e,-}x^{(2)}_j\right)\right]\\
    =& (\Bar{x}^{(1)}-\Bar{x}^{(2)}) \sum\limits_{e\in\mathcal{E}}\left(-\frac{1}{N_1}\psi_j^{e,-}x^{(2)}_j - \frac{1}{N_2}\psi_j^{e,-}x^{(2)}_j
        + \frac{1}{N_1}\psi_i^{e,-}x^{(1)}_i  + \frac{1}{N_2}\psi_i^{e,-}x^{(1)}_i\right)\\
    =& (\Bar{x}^{(1)}-\Bar{x}^{(2)}) \sum\limits_{e\in\mathcal{E}}\left(\right.-\frac{1}{N_1}\psi_j^{e,-}(\Bar{x}^{(2)}+\hat{x}^{(2)}_j) - \frac{1}{N_2}\psi_j^{e,-}(\Bar{x}^{(2)}+\hat{x}^{(2)}_j)
        + \frac{1}{N_1}\psi_i^{e,-}(\Bar{x}^{(1)}+\hat{x}^{(1)}_i) \\
        &\qquad + \frac{1}{N_2}\psi_i^{e,-}(\Bar{x}^{(1)}+\hat{x}^{(1)}_j)\left.\right)\\
    =& (\Bar{x}^{(1)}-\Bar{x}^{(2)})\left\{ 
        \left[-\left(\frac{1}{N_1}+\frac{1}{N_2}\right)\sum\limits_{e\in\mathcal{E}}\sum\limits_{j\in e}\psi_j^{e,-}\right]\Bar{x}^{(2)}
        + \left[\left(\frac{1}{N_1}+\frac{1}{N_2}\right)\sum\limits_{e\in\mathcal{E}}\sum\limits_{i\in e}\psi_i^{e,-}\right]\Bar{x}^{(1)}\right.\\
        &\qquad \left.\left[-\left(\frac{1}{N_1}+\frac{1}{N_2}\right)\sum\limits_{e\in\mathcal{E}}\sum\limits_{j\in e}\psi_j^{e,-}\right]\hat{x}^{(2)}_j
        + \left[\left(\frac{1}{N_1}+\frac{1}{N_2}\right)\sum\limits_{e\in\mathcal{E}}\sum\limits_{i\in e}\psi_i^{e,-}\right]\hat{x}^{(1)}_i
          \right\}\\
    =& \frac{2}{N_0}(\Bar{x}^{(1)}-\Bar{x}^{(2)})\left[ \sum\limits_{e\in\mathcal{E}}\sum\limits_{i\in e}\psi_i^{e,-}\Bar{x}^{(1)} 
        - \sum\limits_{e\in\mathcal{E}}\sum\limits_{j\in e}\psi_j^{e,-}\Bar{x}^{(2)} \right]\\
        &\qquad + \frac{2}{N_0}(\Bar{x}^{(1)}-\Bar{x}^{(2)})\left[ \sum\limits_{e\in\mathcal{E}}\sum\limits_{i\in e}\psi_i^{e,-}\hat{x}^{(1)}_i 
        - \sum\limits_{e\in\mathcal{E}}\sum\limits_{j\in e}\psi_j^{e,-}\hat{x}^{(2)}_j \right].
\end{align}
% \end{equation}
We denote 
$$\textrm{Pse}(\hat{x}^{(1)},\hat{x}^{(2)}):= \sum\limits_{e\in\mathcal{E}}\sum\limits_{i\in e}\psi_i^{e,-}\hat{x}^{(1)}_i 
        - \sum\limits_{e\in\mathcal{E}}\sum\limits_{j\in e}\psi_j^{e,-}\hat{x}^{(2)}_j.$$
and 
    $$\textrm{Pse}(\Bar{x}^{(1)},\Bar{x}^{(2)}):= \sum\limits_{e\in\mathcal{E}}\sum\limits_{i\in e}\psi_i^{e,-}\Bar{x}^{(1)} 
        - \sum\limits_{e\in\mathcal{E}}\sum\limits_{j\in e}\psi_j^{e,-}\Bar{x}^{(2)}.$$
Assume there exist constants $c_m,c_v,$ such that
\begin{equation}
    \textrm{Pes}(\Bar{x}^{(1)},\Bar{x}^{(2)}) \ge c_m  (\Bar{x}^{(1)} - \Bar{x}^{(2)}).
\end{equation}\label{cond:cm}
Then, by Cauchy's inequality, for any $c_1,$ we have
\begin{equation}
    \begin{aligned}
        &\frac{1}{2}\frac{\mathrm{d}}{\mathrm{d}t}\|\Bar{x}^{(1)}-\Bar{x}^{(2)}\|^2 \\
        =& \frac{2}{N_0}\textrm{Pes}(\Bar{x}^{(1)},\Bar{x}^{(2)})(\Bar{x}^{(1)}-\Bar{x}^{(2)}) + \frac{2}{N_0}\textrm{Pes}(\hat{x}^{(1)},\hat{x}^{(2)})(\Bar{x}^{(1)}-\Bar{x}^{(2)})\\
        \ge & \left(\frac{2c_m}{N_0} - c_1\right) \|\Bar{x}^{(1)}-\Bar{x}^{(2)}\|^2 + c_1 \|\Bar{x}^{(1)}-\Bar{x}^{(2)}\|^2 + \frac{2}{N_0}\textrm{Pes}(\hat{x}^{(1)},\hat{x}^{(2)})(\Bar{x}^{(1)}-\Bar{x}^{(2)})\\
        \ge & \left(\frac{2c_m}{N_0} - c_1\right) \|\Bar{x}^{(1)}-\Bar{x}^{(2)}\|^2 
            - \frac{1}{c_1}\frac{2}{N_0}\left(\textrm{Pes}(\hat{x}^{(1)},\hat{x}^{(2)})\right)^2\\
        \ge & \left(\frac{2c_m}{N_0} - c_1\right) \|\Bar{x}^{(1)}-\Bar{x}^{(2)}\|^2 
            - \frac{4}{c_1 N_0}\sum\limits_{e\in\mathcal{E}}\left[
            \left(\sum\limits_{i\in e} \psi_i^{e,-}\hat{x}^{(1)}_i\right)^2
            + \left(\sum\limits_{j\in e} \psi_j^{e,-}\hat{x}^{(2)}_j\right)^2
            \right]\\
        \ge & \left(\frac{2c_m}{N_0} - c_1\right) \|\Bar{x}^{(1)}-\Bar{x}^{(2)}\|^2 
            - \frac{4}{c_1 N_0}\sum\limits_{e\in\mathcal{E}}\left[
            \|\psi^{e,-}\|^2\sum\limits_{i\in e}(\hat{x}^{(1)}_i)^2
            + \|\psi^{e,-}\|^2\sum\limits_{j\in e}(\hat{x}^{(2)}_j)^2
            \right]\\
        \ge & \left(\frac{2c_m}{N_0} - c_1\right) \|\Bar{x}^{(1)}-\Bar{x}^{(2)}\|^2  
            - \frac{4(D_2^-)^2}{c_1 N_0}\sum\limits_{e\in\mathcal{E}}\left[
            \sum\limits_{i\in e}(\hat{x}^{(1)}_i)^2 + \sum\limits_{j\in e}(\hat{x}^{(2)}_j)^2
            \right]\\
        \ge & \left(\frac{2c_m}{N_0} - c_1\right) \|\Bar{x}^{(1)}-\Bar{x}^{(2)}\|^2 
            - \frac{4(D_2^-)^2}{c_1 N_0}k [\widehat{M_2}].  \\
    \end{aligned}
\end{equation}
\bs

\begin{lemma}\label{lem: L2 estimate for hatM2}
Let $u,v$ be the solution to \eqref{eq: bi}. Then $\widehat{M}_2$ satisfies
    \begin{equation}
    \frac{1}{2}\frac{\mathrm{d}}{\mathrm{d}t}\widehat{M}_2\le C_2 \widehat{M}_2 + 2c_2\|\Bar{x}^{(1)}-\Bar{x}^{(2)}\|^2,
\end{equation}
where 
\begin{equation}
    C_2:= -k\left(C^A - \frac{D^-}{2} + \frac{(D^-)^2}{4c_2} - \frac{\delta}{k}\right),
\end{equation}
and $c_2$ is an arbitrary positive constant.
\end{lemma}

\textit{Proof.}
Subtracting \eqref{eq:4.7 diff of mean} from \eqref{eq: bi} gives $\Dot{\hat{x}}^{(1)}_i$. Then we have
\begin{equation}
\allowdisplaybreaks
    \begin{aligned}
        &\frac{1}{2}\frac{\mathrm{d}}{\mathrm{d}t}\left(\frac{1}{N_1}\sum\limits_{i=1}^{N_1}\|\hat{x}^{(1)}_i\|^2\right) \\
        =& \frac{1}{N_1}\sum\limits_{i=1}^{N_1}\hat{x}^{(1)}_i\Dot{\hat{x}}^{(1)}_i\\
        =&  \frac{1}{N_1}\sum\limits_{i=1}^{N_1}\hat{x}^{(1)}_i \sum\limits_{e\in\mathcal{E}(i)}
            \left[ \sum\limits_{i^\prime \in e}h_{i,i^\prime}^{e,+} (x^{(1)}_{i^\prime}-x^{(1)}_i) 
            -  \sum\limits_{j\in e}h_{i,j}^{e,-} (x^{(2)}_j-x^{(1)}_i) \right] \\
            &\qquad +\frac{\delta}{N_1}\sum\limits_{i=1}^{N_1}\hat{x}^{(1)}_i x^{(1)}_i(1-(x^{(1)}_i)^2)\\
        =&  \frac{1}{N_1}\sum\limits_{i=1}^{N_1}\hat{x}^{(1)}_i \sum\limits_{e\in\mathcal{E}(i)}
            \left[ \sum\limits_{i^\prime \in e}h_{i,i^\prime}^{e,+} (\hat{x}^{(1)}_{i^\prime}-\hat{x}^{(1)}_i) 
            -  \sum\limits_{j\in e}h_{i,j}^{e,-} (x^{(2)}_j-x^{(1)}_i) \right] \\
            &\qquad +\frac{\delta}{N_1}\sum\limits_{i=1}^{N_1}\hat{x}^{(1)}_i x^{(1)}_i(1-(x^{(1)}_i)^2)\\
        =&  \frac{1}{N_1}\sum\limits_{i=1}^{N_1}\hat{x}^{(1)}_i \sum\limits_{e\in\mathcal{E}(i)}
            \sum\limits_{i^\prime \in e}h_{i,i^\prime}^{e,+} (\hat{x}^{(1)}_{i^\prime}-\hat{x}^{(1)}_i) 
            - \frac{1}{N_1}\sum\limits_{i=1}^{N_1}\hat{x}^{(1)}_i \sum\limits_{e\in\mathcal{E}(i)} 
            \sum\limits_{j\in e}h_{i,j}^{e,-} (\hat{x}^{(2)}_j-\hat{x}^{(1)}_i)\\
            &\, - \frac{1}{N_1}\sum\limits_{i=1}^{N_1}\hat{x}^{(1)}_i \sum\limits_{e\in\mathcal{E}(i)} 
            \sum\limits_{j\in e}h_{i,j}^{e,-} (\Bar{x}^{(2)}-\Bar{x}^{(1)})
            +\frac{\delta}{N_1}\sum\limits_{i=1}^{N_1}\hat{x}^{(1)}_i x^{(1)}_i(1-(x^{(1)}_i)^2)\\
        =&: I_1 + I_2 + I_3 + I_4.
    \end{aligned}
\end{equation}

$I_1$ can be defined by
\begin{equation}
    \begin{aligned}
        I_1 =& \frac{1}{N_1}\sum\limits_{i=1}^{N_1}\hat{x}^{(1)}_i \sum\limits_{e\in\mathcal{E}(i)}
            \sum\limits_{i^\prime \in e}h_{i,i^\prime}^{e,+} (\hat{x}^{(1)}_{i^\prime}-\hat{x}^{(1)}_i)\\
            =& \frac{1}{N_1}\sum\limits_{e\in\mathcal{E}}
            \sum\limits_{i,i^\prime \in e}h_{i,i^\prime}^{e,+}(\hat{x}^{(1)}_{i^\prime}-\hat{x}^{(1)}_i)\hat{x}^{(1)}_i\\
            =& \frac{1}{N_1}\sum\limits_{e\in\mathcal{E}}((\hat{x}^{(1)})^{e})^\top A^e (\hat{x}^{(1)})^{e},
    \end{aligned}
\end{equation}

where $(\hat{x}^{(1)})^{e} := (\hat{x}^{(1)}_{i_{1}},\cdots,\hat{x}^{(1)}_{i_{|e|}})^\top$ and $A^e$ is given by \eqref{eq:matrix3.1} for each $e.$ Thus $I_1$ is bounded by 
\begin{equation}
    \frac{1}{N_1}\sum\limits_{e\in\mathcal{E}}((\hat{x}^{(1)})^{e})^\top A^e (\hat{x}^{(1)})^{e} 
    \le - \frac{1}{N_1}\sum\limits_{e\in\mathcal{E}}C^A \|(\hat{x}^{(1)})^e\|^2
    \le - \frac{c_r}{N_1}\sum\limits_{i=1}^{N_1}C^A \|\hat{x}^{(1)}_i\|^2,
\end{equation}
where $c_r$ is a constant larger than $1$ related to the repetition of $\{\hat{x}^{(1)}_i\}$ in all hyperedges.

$I_2$ can be controlled by
% \begin{equation}
    \begin{align}
        I_2 =& - \frac{1}{N_1}\sum\limits_{i=1}^{N_1}\hat{x}^{(1)}_i \sum\limits_{e\in\mathcal{E}(i)} 
            \sum\limits_{j\in e}h_{i,j}^{e,-} (\hat{x}^{(2)}_j-\hat{x}^{(1)}_i)\\
            =& -\frac{1}{N_1}\sum\limits_{e\in\mathcal{E}}\sum\limits_{i,j\in e} h_{i,j}^{e,-}(\hat{x}^{(2)}_j-\hat{x}^{(1)}_i)\hat{x}^{(1)}_i\\
            =& -\frac{1}{N_1}\sum\limits_{e\in\mathcal{E}}\sum\limits_{i,j\in e} h_{i,j}^{e,-}\hat{x}^{(2)}_j\hat{x}^{(1)}_i
                + \frac{1}{N_1}\sum\limits_{e\in\mathcal{E}}\sum\limits_{i,j\in e} h_{i,j}^{e,-}\|\hat{x}^{(1)}_i\|^2\\
            \le& \frac{1}{N_1}\sum\limits_{e\in\mathcal{E}}\sum\limits_{i,j\in e} h_{i,j}^{e,-}\frac{1}{2}(\|\hat{x}^{(2)}_j\|^2+\|\hat{x}^{(1)}_i\|^2)
                + \frac{1}{N_1}\sum\limits_{e\in\mathcal{E}}D_e^-\sum\limits_{i\in e}\|\hat{x}^{(1)}_i\|^2\\
            \le& \frac{1}{2N_1}\sum\limits_{e\in\mathcal{E}}\sum\limits_{j\in e}\psi_j^{e,-}\|\hat{x}^{(2)}_j\|^2 
                + \frac{3D^-}{2N_1}\sum\limits_{e\in\mathcal{E}}\sum\limits_{i\in e}\|\hat{x}^{(1)}_j\|^2 \\
            \le& \frac{D^-}{2N_1}\sum\limits_{e\in\mathcal{E}}\sum\limits_{j\in e}\|\hat{x}^{(2)}_j\|^2 
                + \frac{3D^-}{2N_1}\sum\limits_{e\in\mathcal{E}}\sum\limits_{i\in e}\|\hat{x}^{(1)}_j\|^2 \\
            \le& \frac{D^- k}{2N_1}\sum\limits_{j=1}^{N_2}\|\hat{x}^{(2)}_j\|^2 
                + \frac{3D^- k}{2N_1}\sum\limits_{i=1}^{N_1}\|\hat{x}^{(1)}_i\|^2. 
    \end{align}
% \end{equation}

$I_3$ has the below estimate for any constant $c_2>0$:
\begin{equation}
    \begin{aligned}
        I_3 =& - \frac{1}{N_1}\sum\limits_{i=1}^{N_1}\hat{x}^{(1)}_i \sum\limits_{e\in\mathcal{E}(i)} 
            \sum\limits_{j\in e}h_{i,j}^{e,-} (\Bar{x}^{(2)}-\Bar{x}^{(1)})\\
            \le& c_2\|\Bar{x}^{(1)}-\Bar{x}^{(2)}\|^2 + \frac{1}{4c_2N_1}\sum\limits_{e\in\mathcal{E}}\sum\limits_{i,j\in e}\|h_{i,j}^{e,-}\|^2\|\hat{x}^{(1)}_i\|^2\\
            \le& c_2\|\Bar{x}^{(1)}-\Bar{x}^{(2)}\|^2 + \frac{(D^-)^2}{4c_2N_1}\sum\limits_{e\in\mathcal{E}}\sum\limits_{i,j\in e}\|\hat{x}^{(1)}_i\|^2\\
            \le& c_2\|\Bar{x}^{(1)}-\Bar{x}^{(2)}\|^2 + \frac{(D^-)^2 k}{4c_2N_1}\sum\limits_{e\in\mathcal{E}}\sum\limits_{i=1}^{N_1}\|\hat{x}^{(1)}_i\|^2.
    \end{aligned}
\end{equation}
Define
\begin{equation}
    \begin{aligned}
    I_4 :=&\frac{1}{N_1}\delta\sum\limits_{i=1}^{N_1}\hat{x}^{(1)}_i x^{(1)}_i (1-\|x^{(1)}_i\|^2) \\
    =& \frac{\delta}{N_1}\sum\limits_{i=1}^{N_1}\hat{x}^{(1)}_i(\hat{x}^{(1)}_i + \Bar{x}^{(1)})(1-(x^{(1)}_i)^2)\\
        =& \frac{\delta}{N_1}\sum\limits_{i=1}^{N_1}(\hat{x}^{(1)}_i)^2 
            - \frac{\delta}{N_1}\sum\limits_{i=1}^{N_1}(\hat{x}^{(1)}_i)^2(x^{(1)}_i)^2
            +\frac{\delta}{N_1}\sum\limits_{i=1}^{N_1}\hat{x}^{(1)}_i\Bar{x}^{(1)}
            -\frac{\delta}{N_1}\sum\limits_{i=1}^{N_1}\hat{x}^{(1)}_i (x^{(1)}_i)^2 \Bar{x}^{(1)}\\
        =& \frac{\delta}{N_1}\sum\limits_{i=1}^{N_1}(\hat{x}^{(1)}_i)^2 
            - \frac{\delta}{N_1}\sum\limits_{i=1}^{N_1}\hat{x}^{(1)}_i \|x^{(1)}_i\|^2 x^{(1)}_i.
    \end{aligned}
\end{equation}
Note that
\begin{equation}
    \begin{aligned}
        \sum\limits_{i=1}^{N_1}\hat{x}^{(1)}_i \|x^{(1)}_i\|^2 x^{(1)}_i
         =& \sum\limits_{i=1}^{N_1}\|x^{(1)}_i\|^2 (\|x^{(1)}_i\|^2 - x^{(1)}_i \Bar{x}^{(1)})\\
         \ge& \frac{1}{2}\sum\limits_{i=1}^{N_1}\|x^{(1)}_i\|^2 (\|x^{(1)}_i\|^2 - \|\Bar{x}^{(1)}\|^2)\\
         =& \frac{1}{2} \sum\limits_{i=1}^{N_1}\|x^{(1)}_i\|^4 -\frac{1}{2}\sum\limits_{i=1}^{N_1}\|x^{(1)}_i\|^2 \|\Bar{x}^{(1)}\|^2\\
         \ge& \frac{1}{2} \sum\limits_{i=1}^{N_1}\|x^{(1)}_i\|^4 -\frac{1}{2N_1}\sum\limits_{i=1}^{N_1}(\|x^{(1)}_i\|^2)^2\\
         \ge& 0.
    \end{aligned}
\end{equation}
Hence,
\begin{equation}
    I_4\le \delta \widehat{M}_2(u).
\end{equation}

Then
\begin{equation}
    \begin{aligned}
        &\frac{\mathrm{d}}{\mathrm{d}t}\left(\frac{1}{2N_1}\sum\limits_{i=1}^{N_1}\|\hat{x}^{(1)}_i\|^2\right)\\
        &\, \le k\left(-\frac{C^A}{N_1} + \frac{3D^-}{2N_1} + \frac{(D^-)^2}{4c_2 N_1}\right)\sum\limits_{i=1}^{N_1}\|\hat{x}^{(1)}_i\|^2 
            + k\frac{D^-}{2N_2}\sum\limits_{j=1}^{N_2}\|\hat{x}^{(2)}_j\|^2 + c_2\|\Bar{x}^{(1)}-\Bar{x}^{(2)}\|^2\\
        &\, \le \left[k\left(-C^A + \frac{3D^-}{2} + \frac{(D^-)^2}{4c_2}\right)+\delta\right]\widehat{M}_2(u)
            + k\frac{D^-}{2}\widehat{M}_2(v) + c_2\|\Bar{x}^{(1)}-\Bar{x}^{(2)}\|^2.
    \end{aligned}
\end{equation}

Similarly,
\begin{equation}
    \begin{aligned}
        &\frac{\mathrm{d}}{\mathrm{d}t}\left(\frac{1}{2N_2}\sum\limits_{j=1}^{N_2}\|\hat{x}^{(2)}_j\|^2\right)\\
        &\, \le \left[k\left(-C^A + \frac{3D^-}{2} + \frac{(D^-)^2}{4c_2}\right)+\delta\right]\widehat{M}_2(v)
            + k\frac{D^-}{2}\widehat{M}_2(u) + c_2\|\Bar{x}^{(1)}-\Bar{x}^{(2)}\|^2.
    \end{aligned}
\end{equation}

Summing them together gives
\begin{equation}
    \frac{1}{2}\frac{\mathrm{d}}{\mathrm{d}t}(\widehat{M_2})\le 
    -k\left(C^A - \frac{D^-}{2} + \frac{(D^-)^2}{4c_2} - \frac{\delta}{k}\right)(\widehat{M_2}) + 2c_2\|\Bar{x}^{(1)}-\Bar{x}^{(2)}\|^2.
\end{equation}
This gives an exponential growth estimate or $\widehat{M}_2$ up to an error term of $\|\Bar{x}^{(1)}-\Bar{x}^{(2)}\|^2.$ 
\bs

\begin{proposition}[$L_2$ separation of \textsf{HAMP-I}]
For \eqref{eq:1st order}, suppose the above assumptions are satisfied.  Define the mean value $\mathbf{\Bar{x}} := \frac{1}{N}\sum\limits_{i=1}^N \mathbf{x}_i$, 
% the deviation value $\mathbf{\hat{x}}_i := \mathbf{x}_i - \mathbf{\Bar{x}}$,
and the second moments $M_2(\mathbf{x}):= \sum \limits_{i=1}^N \mathbf{x}_i^2.$ Then for sufficiently large $N_1,N_2$, there exist constants $\lambda_- , \lambda_+, $ such that if the initial data satisfies 
\begin{equation}
    \lambda(0) := \frac{\widehat{M}_2(0)}{\|\mathbf{\Bar{x}}^{(1)}(0)-\mathbf{\Bar{x}}^{(2)}(0)\|^2} \le \lambda_+,
\end{equation}
then, there holds that  the $L_2$  separation
\begin{equation}
    \lambda(t):= \frac{\widehat{M}_2(t)}{\|\mathbf{\Bar{x}}^{(1)}(t) - \mathbf{\Bar{x}}^{(2)}(t)\|^2}
    \le \lambda_- + (\lambda(0)-\lambda_-)e^{-\mu t},
\end{equation}
with a positive constant $\mu$, where $\widehat{M}_2(t):=M_2(\mathbf{x}^{(1)}(t))+M_2(\mathbf{x}^{(2)}(t))$.
\end{proposition}

\textit{Proof of Proposition~\ref{thm:L2 separation}.}
Lemma \ref{lem: L2 estimate for central dist} gives an exponential growth estimate of $\|\Bar{x}^{(1)}-\Bar{x}^{(2)}\|^2$ up to an error term of $\widehat{M}_2.$ If $N_0$ is large enough, the coefficient of the error term is small.

% \textbf{Step 2:} $L_2$ estimate for $\widehat{M_2}.$\\
Lemma \ref{lem: L2 estimate for hatM2} gives an exponential growth estimate or $\widehat{M_2}$ up to an error term of $\|\Bar{x}^{(1)}-\Bar{x}^{(2)}\|^2.$ If $N_0$ is large enough, the coefficient of the error term is small.

Set 
\begin{equation}\label{cond: delta condition}
    A_{11} := \frac{2c_m}{N_0} - c_1,\, A_{12}:=\frac{4(D_2^-)^2 k}{c_1N_0},\,
    A_{21} = 2c_2,\, A_{22}= k\left(C^A - \frac{D^-}{2} + \frac{(D^-)^2}{4c_2}\right).
\end{equation}
If $N_0$ is large enough, \eqref{cond: delta condition} is satisfied.
\begin{equation}
    (A_{11}+A_{12})^2 - 4A_{21}A_{12}>0.
\end{equation}\label{cond: delta}
Apply Lemma 4.1 in \cite{jin2021collective} then we obtain the $L_2$ separation.
\bs

\begin{remark}
    We can alternatively prove Theorem~\ref{thm:L2 separation} by the method of [\cite{wang2023acmp} Proposition 2 and 3].
\end{remark}

\begin{proposition}[Lower bound of the Dirichlet energy]
     If the hypergraph $\mathcal{H}$ is a connected one, for \eqref{eq:1st order} with the conditions of Theorem \ref{thm:L2 separation}, or for \eqref{eq:2nd order} with conditions of Theorem 5.1 in~\cite{fang2019emergent}, there exists a positive lower bound of the Dirichlet energy.
\end{proposition}

\textit{Proof of Proposition~\ref{prop: lower bound}.}
The relative size between $\|\Bar{x}^{(1)}-\Bar{x}^{(2)}\|^2$  and $\widehat{M_2}$ is an indicator of group separation in the sense of $L_2$: if $\|\Bar{x}^{(1)}-\Bar{x}^{(2)}\|^2$ is much larger than $\widehat{M_2}$, then the two groups are well-separated in average sense. Since the hypergraph is connected, there is a positive bound between different clusters, hence the Dirichlet energy does not decay to zero.

\begin{remark}
The proof for the second-order system \eqref{eq:2nd order} can be proved in a similar way.
\end{remark}

\begin{remark}
The separability of \eqref{eq: 1st order with noise} which is the \eqref{eq:1st order} or \eqref{eq:2nd order} with a stochastic term also holds.
\end{remark}

% \subsection{Results with noise}
% \begin{remark}
% In \eqref{eq: 1st order with noise}, for each finite time $t$, the increment $d B_t$ is a Gaussian white noise characterized by mean zero and variance $\mathbf{I}_d dt$. By the strong law of large numbers, the noise term converges to zero almost surely as $N \to \infty$.
% \end{remark}

% \subsection{Results on second-order system}

\section{Experiment Details}\label{ex_details}

\subsection{Dataset Details}

We selected the most common hypergraph benchmark datasets, and the statistics of nine datasets across different domains are summarized in \tabref{tab:datasets}. The key statistics include the number of nodes, hyperedges, features, classes, average node degree $d_v$, average hyperedge size $|\mathcal{E}|$, and CE homophily. These variations highlight the diversity in dataset scales and structural patterns, which may influence model performance in hypergraph-related tasks. 
\begin{table*}[!thbp]
    \caption{The summary of data statistics.}
    \label{tab:datasets}
    \begin{center}
    \small
    \begin{tabular}{lccccccc}
    \toprule
    Dataset	&\# nodes  & \# hyperedges  &\# features &\# classes  &avg. $d_v$  &avg. $|\mathcal{E}|$  &CE homophily\\
    \midrule
    Cora     &2708  &1579  &1433 &7 &1.767    &3.03   &0.897   \\
    Citeseer &3312  &1079  &3703 &6 &1.044    &3.200  &0.893   \\
    Pubmed   &19717 &7963  &500  &3 &1.756    &4.349  &0.952   \\
    Cora-CA  &2708  &1072  &1433 &7 &1.693    &4.277  &0.803   \\
    DBLP-CA  &41302 &22363 &1425 &6 &2.411    &4.452  &0.869   \\
    \midrule
    Congress &1718  &83105 &100  &2 &427.237  &8.656  &0.555   \\
    House    &1290  &340   &100  &2 &9.181    &34.730 &0.509   \\
    Senate   &282   &315   &100  &2 &19.177   &17.168 &0.498   \\
    Walmart  &88860 &69906 &100  &11 &5.184   &6.589  &0.530   \\
    \bottomrule
    \end{tabular}
    \end{center}
\end{table*}

\subsection{Additional Ablation Studies}\label{ab_details}

\paragraph{Impact of Hidden Dimension on \textsf{HAMP-I} and \textsf{HAMP-II}.} 
We explore the tolerance of our model to different hidden dimensions, as shown in \tabref{tab:hid_dim}. For simplicity, we only vary the size of hidden dimension, while other parameters remain fixed. Overall, these results  demonstrate the robustness of our methods with varying hidden dimensions. It is worth noting that high hidden dimension is key to achieving the best performance of \textsf{HAMP}.

\begin{table*}[!thbp]
    \caption{Impact of hidden dimension evaluated on the hypergraph datasets.}
    \label{tab:hid_dim}
    \small
    \begin{center}
        \begin{tabular}{lcccc|cccc}
        \toprule 
         &\multicolumn{4}{c|}{\textsf{HAMP-I}} &\multicolumn{4}{c}{\textsf{HAMP-II}} \\
         \cmidrule(r){2-5} \cmidrule{6-9}
         &128 &256 &512 &1024  &128 &256 &512 &1024 \\
        \midrule
        Cora      &80.19  &80.55  &\textbf{81.18}  &79.85  &79.59  &79.54  &\textbf{80.80} &79.60  \\
        Citeseer  &73.39  &72.89  &\textbf{75.22} &72.40  &73.95 &74.36 &\textbf{75.33} &75.19 \\
        Pubmed    &88.49  &88.85 &\textbf{89.02}  &88.82  & 88.48 & 88.48 & \textbf{89.05} & 88.78 \\
        Senate    &63.24  &65.35  &\textbf{69.44}  &66.62  & 63.52 & 63.10 & \textbf{70.14} & 62.96 \\
        House     &71.21  &70.62  &72.66  &\textbf{72.72}  &68.30  &69.47   &\textbf{72.60}  &71.21     \\
        \bottomrule
        \end{tabular}
    \end{center}
\end{table*}

\paragraph{Impact of Repulsion, Allen-Cahn force, and Noise on \textsf{HAMP-I} and \textsf{HAMP-II}.} 
We summary ablation studies to investigate the individual and combined effects of repulsion $f_\beta^-$, Allen-Cahn force $f_d$, and noise $B_t$ on both \textsf{HAMP-I} and \textsf{HAMP-II}. \tabref{tab:EX_Ablation2} reports the average node classification accuracy with a standard deviation across seven standard hypergraph benchmarks over 10 runs. Models with the repulsion term enabled outperform their counterparts in some dataset, indicating enhanced ability to distinguish node representations in complex hypergraph structures. The synergy between the repulsion and Allen-Cahn terms further boosts performance, confirming that these particle system-inspired mechanisms play complementary roles. Overall, these improvement confirm the validity of the \textsf{HAMP} construction and further highlight the significant advantages of incorporating particle system theory into the hypergraph message passing learning process.

\begin{table*}[!thbp]
    \caption{Ablation studies on some standard hypergraph benchmarks. The accuracy (\%) is reported with a standard deviation from 10 repetitive runs. (Key: $f_\beta^-$: repulsion; $f_d$: Allen-Cahn; $B_t$: noise.)}
    \label{tab:EX_Ablation2}
    \small
    \begin{center}
    \begin{tabular}{lccc|cccc}
    \toprule
     Homophilic &$f_\beta^-$  &$f_d$  &$B_t$  &Cora  &Citeseer  &Pubmed  &Cora-CA  \\
    \midrule
    \multirow{7}*{\textsf{HAMP-I}}   &\faTimes &\faTimes &\faTimes  
    &76.09$\pm$1.22   &70.53$\pm$1.56  &87.98$\pm$0.38 &83.13$\pm$1.26 \\
       &\faCheck &\faTimes &\faTimes  
    &76.40$\pm$1.56   &70.85$\pm$1.65  &88.25$\pm$0.50 &83.15$\pm$1.36 \\
       &\faTimes &\faCheck &\faTimes 
    &80.31$\pm$1.41   &74.83$\pm$1.70  &88.90$\pm$0.45 &84.77$\pm$1.16 \\
       &\faTimes &\faTimes &\faCheck  
    &75.67$\pm$1.71   &70.59$\pm$1.40  &87.93$\pm$0.52 &82.70$\pm$1.01 \\
      &\faCheck  &\faCheck &\faTimes 
    &80.49$\pm$1.26   &74.96$\pm$1.56  &88.87$\pm$0.40 &85.21$\pm$1.49 \\
      &\faTimes  &\faCheck &\faCheck 
    &80.59$\pm$1.25  &74.67$\pm$1.69   &88.77$\pm$0.44 &84.59$\pm$1.03 \\
      &\faCheck  &\faCheck &\faCheck 
    &\textbf{81.18$\pm$1.30}  &\textbf{75.22$\pm$1.62} &\textbf{89.02$\pm$0.49}  &\textbf{85.23$\pm$1.15}   \\
    \midrule
    \multirow{7}*{\textsf{HAMP-II}}  
       &\faTimes &\faTimes &\faTimes  
    &77.42$\pm$1.44   &71.50$\pm$1.49  &88.68$\pm$0.62   &83.60$\pm$1.45   \\
       &\faCheck &\faTimes &\faTimes  
    &77.08$\pm$1.73   &72.20$\pm$1.14  &88.59$\pm$0.50   &82.95$\pm$1.35    \\
       &\faTimes &\faCheck &\faTimes 
    &80.13$\pm$1.26   &74.37$\pm$1.59  &88.86$\pm$0.55   &84.37$\pm$1.45    \\
       &\faTimes &\faTimes &\faCheck  
    &77.18$\pm$1.61   &71.75$\pm$1.68  &88.68$\pm$0.59   &82.91$\pm$1.28    \\
      &\faCheck  &\faCheck &\faTimes 
    &79.70$\pm$1.36   &73.99$\pm$1.75  &88.82$\pm$0.48   &84.30$\pm$1.32    \\
      &\faTimes &\faCheck &\faCheck 
    &79.50$\pm$1.25   &74.25$\pm$1.28  &88.80$\pm$0.40   &83.31$\pm$1.44    \\
      &\faCheck &\faCheck &\faCheck
    &\textbf{80.80$\pm$1.62}   &\textbf{75.33$\pm$1.61}  &\textbf{89.05$\pm$0.41}   &\textbf{84.89$\pm$1.14}   \\
    \bottomrule
    \end{tabular}
    \end{center}
\end{table*}

\begin{table*}[!thbp]
    \caption{Node Classification on standard hypergraph benchmarks. The accuracy (\%) is reported with a standard deviation from 10 repetitive runs. (Key: $f_\beta^-$: repulsion; $f_d$: Allen-Cahn; $B_t$: noise.)}
    \label{tab:EX_Ablation_2}
    \begin{center}
    \small
    \begin{tabular}{lccc|cccc}
    \toprule
     Heterophilic  &$f_\beta^-$ &$f_d$ &$B_t$  &Congress &Senate &Walmart  &House  \\
    \midrule
    \multirow{7}*{\textsf{HAMP-I}}   &\faTimes &\faTimes &\faTimes  
    &93.51$\pm$1.08  &60.70$\pm$8.38   &69.64$\pm$0.35 &70.50$\pm$1.45 \\
       &\faCheck &\faTimes &\faTimes  
    &93.70$\pm$1.02  &60.00$\pm$8.66   &69.80$\pm$0.45 &70.56$\pm$1.96  \\
      &\faTimes  &\faCheck&\faTimes 
    &94.79$\pm$1.14  &66.76$\pm$5.44   &69.77$\pm$0.28 &71.55$\pm$1.53  \\
       &\faTimes &\faTimes &\faCheck  
    &93.47$\pm$1.13  &60.14$\pm$7.54   &69.86$\pm$0.35 &69.88$\pm$2.48 \\
      &\faCheck  &\faCheck &\faTimes 
    &94.58$\pm$1.25   &67.75$\pm$8.82  &69.76$\pm$0.37 &71.58$\pm$1.87 \\
      &\faTimes &\faCheck &\faCheck 
    &94.67$\pm$1.02   &65.63$\pm$2.98  &69.73$\pm$0.49 &71.55$\pm$2.58 \\
     &\faCheck  &\faCheck &\faCheck  
    &\textbf{95.09$\pm$0.79}  &\textbf{69.44$\pm$6.09}  &\textbf{69.90$\pm$0.38} &\textbf{72.72$\pm$1.77}  \\
    \midrule
    \multirow{7}*{\textsf{HAMP-II}}   
       &\faTimes &\faTimes &\faTimes  
    &94.79$\pm$0.73  &58.73$\pm$7.03 &69.84$\pm$0.25 &69.85$\pm$1.61  \\
       &\faCheck &\faTimes &\faTimes  
    &94.02$\pm$1.10  &61.55$\pm$5.98 &69.91$\pm$0.30 &69.35$\pm$1.87  \\
      &\faTimes  &\faCheck  &\faTimes 
    &94.19$\pm$1.07  &62.82$\pm$6.44 &69.89$\pm$0.31 &70.96$\pm$2.06   \\
       &\faTimes &\faTimes &\faCheck  
    &94.35$\pm$1.14  &60.14$\pm$5.10 &69.84$\pm$0.37 &70.46$\pm$2.08  \\
      &\faCheck  &\faCheck  &\faTimes 
    &94.58$\pm$0.86   &61.97$\pm$8.42 &69.92$\pm$0.33 &71.36$\pm$1.68 \\
      &\faTimes   &\faCheck  &\faCheck  
    &94.12$\pm$0.63  &64.51$\pm$4.19  &69.86$\pm$0.34  &70.96$\pm$2.76 \\
      &\faCheck  &\faCheck &\faCheck 
    &\textbf{95.26$\pm$1.34}   &\textbf{70.14$\pm$6.08} &\textbf{69.94$\pm$0.37} &\textbf{72.60$\pm$1.23} \\
    \bottomrule
    \end{tabular}
    \end{center}
\end{table*}

\subsection{Time-Memory Tradeoff Analysis}\label{time_memory}
\begin{figure}[!htbp]
\centering
\includegraphics[width=0.65\textwidth]{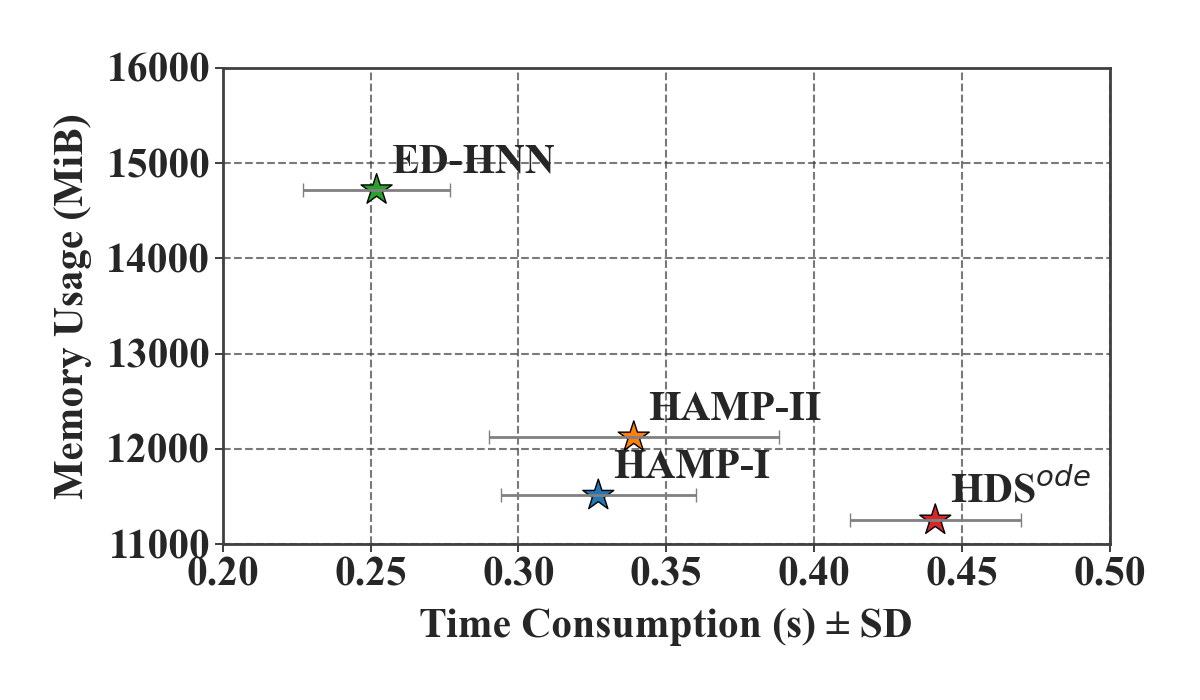}
\caption{Time-Memory tradeoff analysis of different methods on Walmart dataset. SD denotes the standard deviation of time consumption.}
\label{fig:time}
\end{figure}
We intuitively reveal the differences of different methods with a single-layer network in the time-memory trade-off on Walmart dataset. As shown in \figref{fig:time}, \textsf{HAMP-I} and \textsf{HAMP-II} methods demonstrate a notable trade-off between time efficiency and memory consumption. The experimental results reveal: 
\begin{itemize}
    \item Memory usage: \textsf{HAMP-I} and \textsf{HAMP-II} maintain memory consumption within 11000-12000 MiB, achieving a 20-26\% reduction compared to ED-HNN, while being comparable to HDS$^{ode}$.
    \item Time efficiency: Although the time consumption for \textsf{HAMP-I} and \textsf{HAMP-II} runtime slightly exceeds ED-HNN (0.1s), it outperforms the baseline HDS$^{ode}$.
\end{itemize}

\subsection{Hyperparameters}\label{hyperpa}
To ensure fairness, we follow the same training recipe as ED-HNN.
Specifically, we train the model for 500 epochs using the Adam optimizer with the learning rate of 0.001 and no weight decay during the training phases. And we apply early stopping with a patience of 50. For the stability, we run 10 trials with different seed and report the results of mean and the standard deviation. All experiments are implemented on an NVIDIA RTX 4090 GPU with Pytorch.

We explore the parameter space by grid search, where the search ranges for each critical hyperparameter are delineated below: 
\begin{itemize}
    \item Dropout rate in \{0.1, 0.2, 0.3, 0.4, 0.5, 0.6, 0.7, 0.8, 0.9\};
    \item Layer number of classifier in \{1, 2, 3\}; 
    \item The hidden dimension of classifier in \{128, 256, 512\}; 
    \item Hidden dimension of model in \{128, 256, 512, 1024\}; 
    \item step size of solver in \{0.09, 0.1, 0.15, 0.2, 0.25\};
    \item $\gamma$ of repulsive force in \{0.01, 0.02, 0.03, 0.04, 0.05, 0.06, 0.07, 0.08, 0.09, 0.1, 0.11, 0.12, 0.13, 0.14, 0.15\};
    \item Initial values of learnable parameters $\delta$ of damping term in \{0, 1, 2, 3, 4, 5, 6, 7, 8, 9, 10, 11, 12, 13, 14, 15\};
    \item Initial values of learnable parameters $\epsilon$ of noise term in \{0, 0.1, 0.3\};
\end{itemize}

\tabref{tab:Hyperparameters_mp1} and \tabref{tab:Hyperparameters_mp2} summarize the best hyperparameters on standard hypergraph benchmarks using \textsf{HAMP-I} and \textsf{HAMP-II}, respectively. For fairness, a linear layer is added to perform feature mapping when conducting HDS$^{ode}$ experiments. The optimal hyperparameters for node classification on standard hypergraph benchmarks is achieved by the HDS$^{ode}$ algorithm, as demonstrated in \tabref{tab:Hyperparameters_HDS}.

\begin{table*}[!htbp]
    \caption{The best hyperparameters of Node Classification on standard hypergraph benchmarks using the \textsf{HAMP-I} algorithm.}
    \label{tab:Hyperparameters_mp1}
    \begin{center}
    \small
    \begin{tabular}{lcccccccc}
    \toprule
    Dataset	 &model. hd &cls. hd and \# layers  &time &step size &$\delta$ &$\gamma$ &dropout &$\epsilon$ \\
    \midrule
    Cora     &512     &128, 1      &1        &0.1      &12   &0.05   &0.4 &0 \\
    Citeseer &512     &512, 1      &0.6      &0.1      &6    &0.05   &0.2 &0 \\
    Pubmed   &512     &256, 1      &0.2      &0.1     &15   &0.1   &0.5 &0 \\
    Cora-CA  &512     &512, 2      &0.4      &0.2     &4   &0.05   &0.9 &0 \\
    DBLP-CA  &256     &128, 2      &1.1      &0.1     &11   &0.12   &0.2 &0 \\
    Congress &128     &128, 2      &1.4      &0.1      &1    &0.08   &0.3 &0 \\
    House    &1024     &512, 3  &1.05        &0.15     &3   &0.05   &0.8  &0 \\
    Senate   &512     &256, 2      &0.6      &0.1      &10   &0.05   &0.7 &0  \\
    Walmart  &256     &128, 2      &1.75     &0.25     &0    &0.02   &0.3 &0 \\
    \bottomrule
    \end{tabular}
\end{center}
\end{table*}
\begin{table*}[!htbp]
    \caption{The best hyperparameters of Node Classification on standard hypergraph benchmarks using the \textsf{HAMP-II} algorithm.}
    \label{tab:Hyperparameters_mp2}
    \begin{center}
    \small
    \begin{tabular}{lccccccccc}
    \toprule
    Dataset	 &model. hd &cls. hd and layers  &time &step size &$\delta$ &$\gamma$  &dropout &$\epsilon$\\
    \midrule
    Cora     &512    &512, 1    &1.9   &0.1     &5    &0.12   &0.3 &0.1  \\
    Citeseer &512    &512, 1    &1.8   &0.15    &8    &0.13  &0.6 &0  \\
    Pubmed  &512     &256, 1    &0.6   &0.09    &5    &0.09  &0.3 &0 \\
    Cora-CA  &512    &128, 2    &0.75   &0.25     &3     &0.01   &0.7 &0 \\
    DBLP-CA  &256    &128, 2    &3.45  &0.15    &7     &0.09  &0.3  &0  \\
    Congress &128    &128, 2    &6.25  &0.25    &0     &0.01  &0.3 &0 \\
    House    &512    &256, 2    &1.6   &0.1     &8     &0.14  &0.8  &0  \\
    Senate   &512    &256, 2    &3     &0.2     &13    &0.05  &0.3 &0.3 \\
    Walmart  &256    &128, 2    &2.5   &0.25    &0     &0.02  &0.3 &0 \\
    \bottomrule
    \end{tabular}
\end{center}
\end{table*}

\begin{table*}[!htbp]
    \caption{The best hyperparameters of Node Classification on standard hypergraph benchmarks using the HDS$^{ode}$ algorithm.}
    \label{tab:Hyperparameters_HDS}
    \begin{center}
    \small
    \begin{tabular}{lcccccc}
    \toprule
    Dataset	 &model. hd &cls. hd and layers  &\# layer model. &alpha$_v$  &alpha$_e$ &step  \\
    \midrule
    Cora     &512     &256, 2      &10     &0.05    &0.9       &20  \\
    Citeseer &512     &512, 1      &5      &0.05    &0.9       &20   \\
    Pubmed   &512     &256, 1      &12     &0.05    &0.9       &20  \\
    Cora-CA  &512     &512, 2      &9      &0.05    &0.9       &20  \\
    DBLP-CA  &256     &256, 2      &15     &0.05    &0.9       &20  \\
    Congress &256     &128, 2      &7      &0.25    &0.9       &5    \\
    House    &512     &256, 2      &10     &0.05    &0.9       &20  \\
    Senate   &512     &256, 2      &9      &0.05    &0.9       &20   \\
    Walmart  &256     &128, 2      &6      &0.25    &0.9       &5  \\
    \bottomrule
    \end{tabular}
\end{center}
\end{table*}

\end{document}